\documentclass[conference]{IEEEtran}
\IEEEoverridecommandlockouts
\usepackage{cite}
\usepackage{amsmath,amssymb,amsfonts}
\usepackage{subcaption}
\usepackage{algorithmic}
\usepackage{graphicx}
\usepackage{textcomp}
\usepackage{multirow}
\usepackage{xcolor}
\def\BibTeX{{\rm B\kern-.05em{\sc i\kern-.025em b}\kern-.08em
    T\kern-.1667em\lower.7ex\hbox{E}\kern-.125emX}}
\begin{document}

\title{HASeparator: Hyperplane-Assisted Softmax
\thanks{* denotes equal contribution to this work.}
}

\author{\IEEEauthorblockN{Ioannis Kansizoglou\textsuperscript{1 *}, Nicholas Santavas\textsuperscript{1 *}, Loukas Bampis\textsuperscript{1}, and Antonios Gasteratos\textsuperscript{1}}
\\
\IEEEauthorblockN{\textsuperscript{1}\small{\textit{Laboratory of Robotics and Automation}, Democritus University of Thrace, Xanthi, Greece}}
\IEEEauthorblockN{\small{\{ikansizo, nsantava, lbampis, agaster\}@pme.duth.gr}}}

\maketitle

\begin{abstract}
Efficient feature learning with Convolutional Neural Networks (CNNs) constitutes an increasingly imperative property since several challenging tasks of computer vision tend to require cascade schemes and modalities fusion.
Feature learning aims at CNN models capable of extracting embeddings, exhibiting high discrimination among the different classes, as well as intra-class compactness.
In this paper, a novel approach is introduced that \textit{has separator}, which focuses on an effective hyperplane-based segregation of the classes instead of the common class centers separation scheme. Accordingly, an innovatory separator, namely the Hyperplane-Assisted Softmax separator (\textit{HASeparator}), is proposed that demonstrates superior discrimination capabilities, as evaluated on popular image classification benchmarks.
\end{abstract}


\section{Introduction}\label{Intro}
Image recognition with Convolutional Neural Network (CNN) schemes constitutes the approach that nowadays excels in the majority of the computer vision tasks~\cite{liu2017survey,santavas2020attention,arandjelovic2016netvlad}.
Forming one of the most widespread ones, image classification is responsible for dividing the input images into several predefined identities, by exploiting the softmax loss~\cite{rawat2017deep}.
Consequently, with the rising advancements in computer vision, several methods have proposed relevant schemes that learn directly an intermediate representation either for feature fusion in multi-modal tasks~\cite{kansizoglou2019active} or for feature extraction in image retrieval challenges, like face verification~\cite{masi2018deep} and place recognition~\cite{arandjelovic2016netvlad}.
In such cases, the exploitation of softmax is not capable of creating discriminative features since it is not explicitly designed for this task~\cite{deng2019arcface}.

The main challenge of the feature discrimination task focuses on forcing a CNN architecture to learn features that display both high \textit{intra-class compactness} and \textit{inter-class separation}~\cite{deng2019arcface}, indicating small distances between feature vectors of the same class and large distances between vectors of different ones.
Moreover, it is obligatory that the classification performance should not be harmed by the satisfaction of the above criteria.
In this regard, the need for advanced loss functions~\cite{schroff2015facenet,wen2016discriminative,zhang2017range,cai2018island} or additional constraints in softmax loss~\cite{wang2017normface,ranjan2017l2,liu2017sphereface,wang2018cosface,deng2019arcface} has emerged.
To achieve that, the following four approaches have been developed, proposing: \textit{(a)} the insertion of a distance margin between triplets of samples~\cite{schroff2015facenet}, \textit{(b)} the penalization of the feature vector according to its distance from the center of its target class~\cite{wen2016discriminative}, \textit{(c)} the penalization of the different class centers according to their distance or \textit{(d)} the insertion of an additive margin to the geodesic distance between the sample and the class centers~\cite{deng2019arcface}.   

Paying attention to the above instances, one can notice that except for the first case \textit{(a)}, which has already been proved computationally intensive, the rest are based on the approximative notion that weights correspond to the classes' centers~\cite{liu2017sphereface,deng2019arcface}.
Yet, as long as the precise calculation of the actual class centers remains computationally ineffective, the above approximations maintain an inbred error being harmful for discrimination performance.
\begin{figure}[]
    \centering
    \begin{subfigure}[b]{0.49\textwidth}
        \centering
        \includegraphics[width=0.5\textwidth]{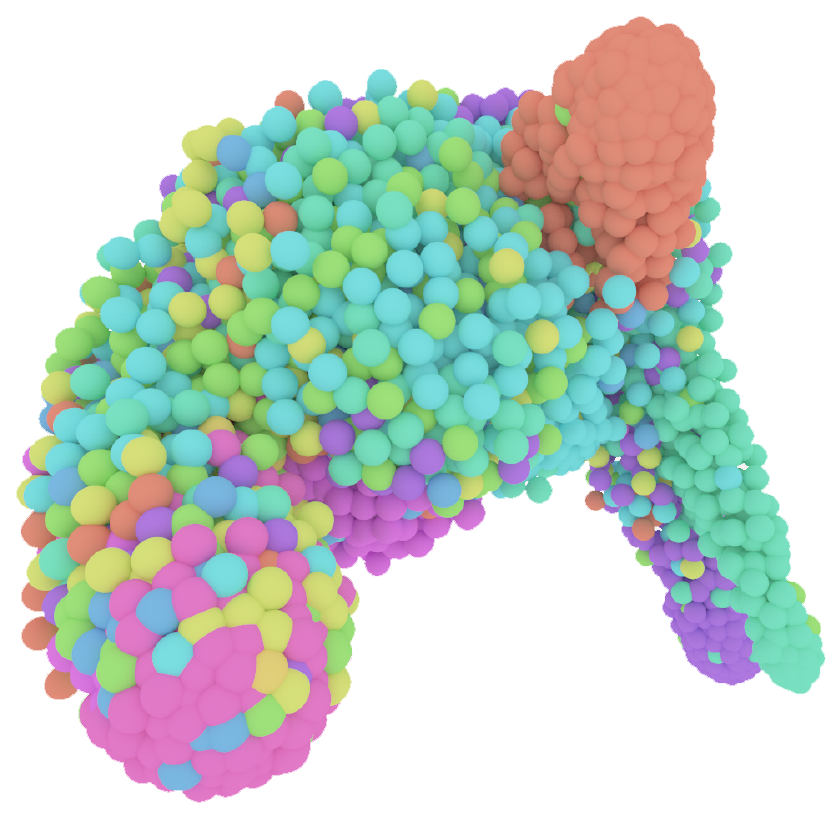}          
    	\label{fig:sub1a}
    \end{subfigure}
    \vskip\baselineskip
    \centering
    \begin{subfigure}[b]{0.49\textwidth}
		\centering    	
        \includegraphics[width=0.6\textwidth]{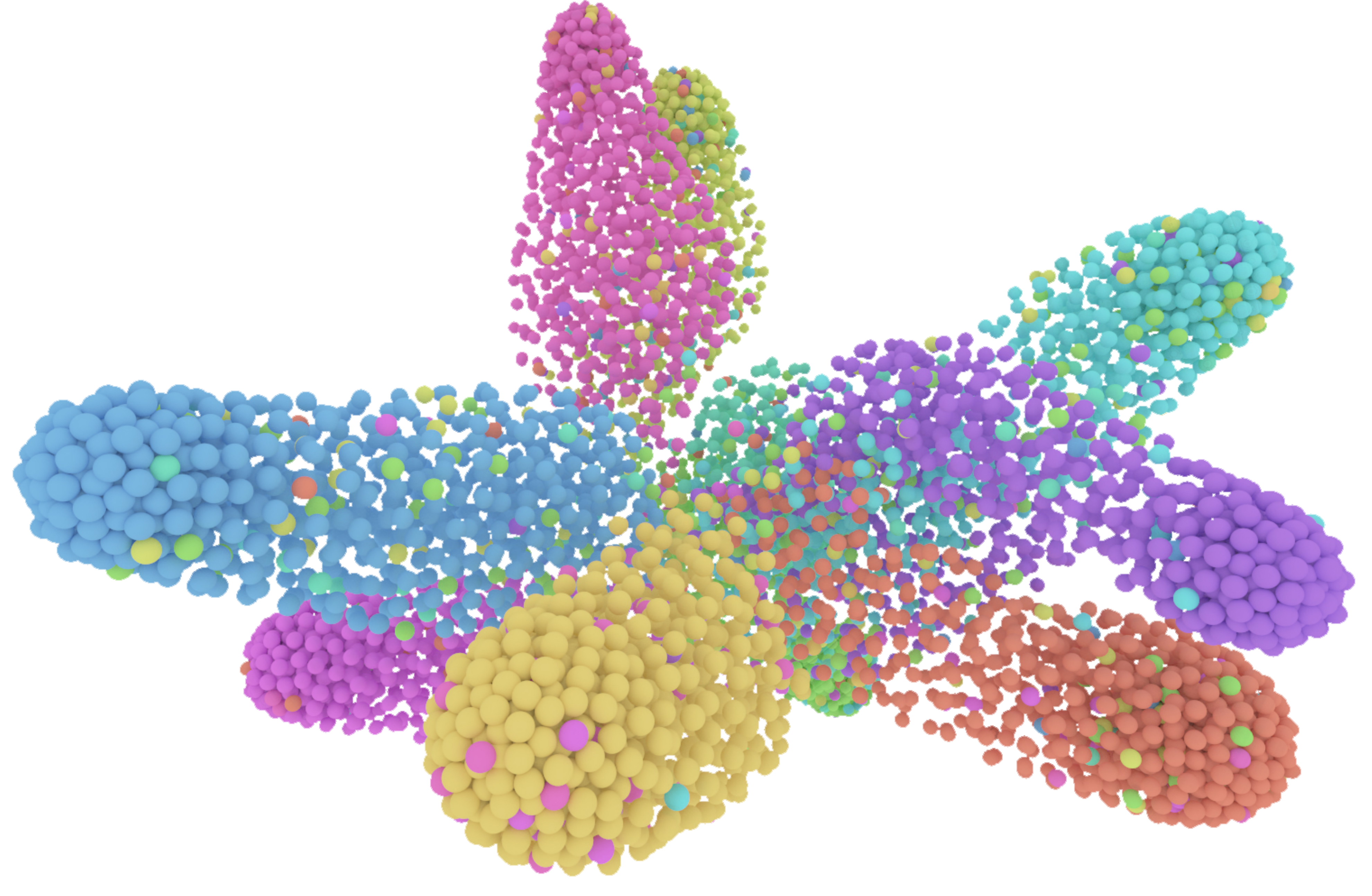}          
    	\label{fig:sub1b}
    \end{subfigure}
    \centering
    \caption[]
    {\small Features learned by \textit{Softmax} (\textit{upper}) and \textit{HASeparator} (\textit{lower}) with ResNet-18 on the testing set of CIFAR-10 visualized on a 3-D space using t-SNE\cite{maaten2008visualizing}.}
    \label{fig:1}
\end{figure}
That have been said, the paper at hand introduces for the first time the usage of an additive penalty based on the angular distance between the embedding and the separation hyperplanes of the target class, which is accurately and easily computed.
An exemplar visualization regarding the feature discrimination capacity of the proposed Hyperplane-Assisted Softmax separator (\textit{HASeparator}) compared against the softmax loss is illustrated in Fig.~\ref{fig:1} using t-SNE~\cite{maaten2008visualizing}.
\textit{HASeparator} is also compared against the most recent approach of \textit{ArcFace} on widespread image classification benchmarks, aiming to demonstrate the credibility of the hyperplanes' approach in this fundamental challenge.
The contributions of this paper can be summarized as follows:
\begin{itemize}
\item feature discrimination without conventions, introducing an additive penalty based on the angular distance between the sample and the decision boundaries of the target class,
\item visual and quantitative evaluation of the features' discrimination capabilities for a given network,
\item detailed comparison of the proposed method's insensitivity to hyperparameters tuning against a well-established, state-of-the-art method (\textit{ArcFace}~\cite{deng2019arcface}).
\end{itemize}

\section{Related Work}\label{RelWork}
Efficient discrimination in feature learning has gained the attention of researchers over the last years, leading to gradual advancements. 
Among many other retrieval-based tasks, face verification has been greatly benefited by this progress~\cite{taigman2014deepface,sun2014deep}, establishing an abundance of methods~\cite{wen2016discriminative,liu2017sphereface,wang2018cosface,deng2019arcface}. 

\subsection{Loss Functions}
In former works a CNN model was trained, in order to learn an embedded representation of the input image~\cite{taigman2014deepface,sun2014deep}. 
Since the inadequacy of softmax loss function in creating discriminative features for open-set face recognition benchmarks~\cite{huang2008labeled,kemelmacher2016megaface} was quickly denoted, \textit{triplet loss}~\cite{schroff2015facenet} was firstly introduced to extract a robust Euclidean space embedding.
However, triplet loss is particularly inefficient in large-scale training datasets, where the total number of possible triplets becomes remarkably high, also leading to the extension of the needed training steps.
Consequently, several following approaches have been developed, so as to enhance the discrimination capabilities of softmax loss~\cite{deng2017marginal,wen2016discriminative,liu2016large,liu2017sphereface,wang2018cosface,deng2019arcface}.
Among them, \textit{center loss}~\cite{wen2016discriminative} was designed to penalise any feature vector on the strength of the norm of its vector difference from the center of the target class.
Thus, a class's compactness is reinforced at the expense of a considerably increasing computational cost, due to the complexity of defining the class center during each iteration step. Inspired by the \textit{triplet} and \textit{center loss}, several variations have been emerged, like \textit{range}~\cite{zhang2017range} and \textit{island loss}~\cite{cai2018island}.

\subsection{Norm-Constrained Approaches}
The concept of normalizing the last layer's weights was initially studied, as an attempt to replace the last inner-product layer with a cosine similarity one~\cite{wang2017normface}, extensively used in open-set challenges of face recognition~\cite{zheng2015triangular}. 
Consequently, the outputs of the specific layer were forced to an $l_2$-constraint as well, achieving improved intra-class compactness~\cite{ranjan2017l2}. The positive effect of feature and weights normalization, led to its adoption by most of the following approaches~\cite{liu2017sphereface,wang2018cosface,deng2019arcface}.

\subsection{Angular-Constrained Approaches}
Additional to norm, several works investigated the effect of applying angular constraints in the feature space. 
Large margin Softmax (\textit{L-Softmax})~\cite{liu2016large} introduced the first idea of a multiplicative angular margin penalization in the loss function, achieving enhanced discrimination.
\textit{L-Softmax} methodology assumes that the weight of each class conceptually approximates the center of the class.
Based on the above assumption, \textit{SphereFace} (\textit{A-Softmax})~\cite{liu2017sphereface} improved state-of-the-art performance, by combining the angular constraints of~\cite{liu2016large} with weight normalization.
However, \textit{SphereFace} was proven unstable during training, requiring the supervision from a softmax loss function.
In order to cope with this problem, \textit{CosFace}~\cite{wang2018cosface} proposed an additive cosine margin that is applied directly to the target logit, yielding to a more complex implementation.
Finally, \textit{ArcFace}~\cite{deng2019arcface} adopted an additive angular margin, succeeding a quite simple feature discrimintation method that outperformed all previous works in the field on several face recognition benchmarks~\cite{huang2008labeled,kemelmacher2016megaface}.

Yet, all the aforementioned methods, employing an angular constraint, assume that the weights of the last fully connected layer account for the classes' centers. 
Inspired by a recent theoretical approach regarding the feature space in this layer~\cite{kansizoglou2020deep}, we argue that the above assumption is not precisely true and, as a result, we argue that this fact reduces the quality of the final solution.
To this end, we propose \textit{HASeparator} that is a clearly different approach without adopting any convention and focusing on the separation hyperplanes between the classes instead of their centers.
Its main advantage lies with the direct definition and straightforward computation of hyperplanes.
\textit{HASeparator} displays a quite simple implementation procedure and competes \mbox{\textit{ArcFace}} in feature discrimination. 

\section{Method}

\subsection{Intuition}\label{Intuit}

The inspiration behind \textit{HASeparator}'s implementation rests upon two main notions.
The first one, already mentioned in Section~\ref{RelWork}, originates from the controversial assumption that the weight vector of each class corresponds to the class center.
Empirically, in order for the above speculation to apply, the weight vectors should bear some kind of symmetry, otherwise the maximum values of softmax do not match with the weight vectors orientations.
This fact, can considerably impact the discrimination performance since it differentiates the orientation of the feature embeddings from the optimal one.
Yet, the separation hyperplane between two classes, \textit{e.g.}, class $1$ and $2$, is simply defined by the characteristic vector $\bar{w}_{12}=\bar{w}_1-\bar{w}_2$~\cite{kansizoglou2020deep}.
Hence, we hypothesize that instead of approximating classes' centers, a method forcing the embeddings to diverge from the classes' separation hyperplanes, banks on a rather solid prospect and should yield superior results.

The most widespread machine learning algorithm with an efficient margin maximization principle for optimal feature discrimination is the Support Vector Machines (SVM)~\cite{cortes1995support}.
To that end, the second main concept of this approach is to utilize and adjust the SVM's properties in the feature space of the last fully connected layer, in order to force both feature embeddings and separation hyperplanes to distance themselves.
Since SVM constitutes a binary classifier, an one-against-all approach is exploited in multi-class challenges.
This procedure can be easily incorporated in a CNN classification layer, as the hyperplanes are defined in a related one-versus-one manner~\cite{kansizoglou2020deep}.

\begin{figure}
    \centering
    \includegraphics[width=0.7\linewidth]{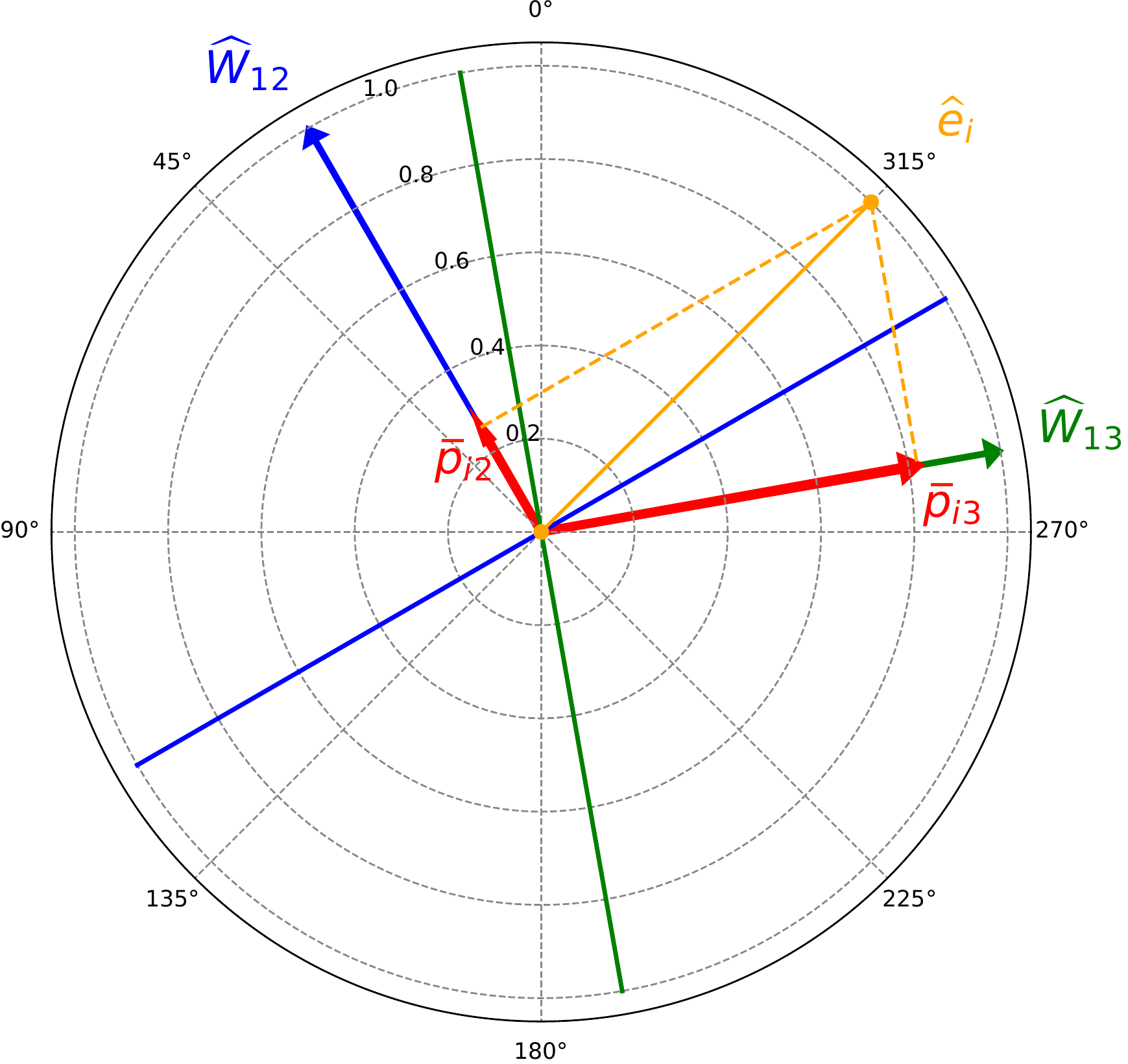}
    \caption{Each unit feature embedding $\hat{e}_i$ is projected on the normal vectors of the target class's separation hyperplanes. By forcing projections' maximization, the embeddings and the hyperplanes learn to diverge.}
        \label{fig:2}
\end{figure}

For convenience, we can think of the 2D feature space depicted in Fig.~\ref{fig:2}.
Moreover, without loss of generality, we assume three target classes.
In this regard, the two one-versus-all separation hyperplanes of class $1$ (between green and blue separation lines) are $\bar{w}_{12}=\bar{w}_1-\bar{w}_2$ and $\bar{w}_{13}=\bar{w}_1-\bar{w}_3$.
Similar to SVM, the feature embedding $\bar{e}_i$ shall maximize its projections on the normal vectors of those lines.
The above vectors are the unit vectors $\hat{w}_{12}=\bar{w}_{12}/ \|\bar{w}_{12}\|$ and $\hat{w}_{13}=\bar{w}_{13}/ \|\bar{w}_{13}\|$.
Note that SVM works with static features.
Thus, only hyperplane optimization is performed and the optimal margin at each iteration is adjusted by the per-class smallest projections (\textit{support vectors}).
However, the features of a CNN are continuously modified during training, adding an extra degree of freedom to our optimization procedure.
Ergo, our method does not use support vectors.
On the contrary, it calculates the projections of all the embeddings from the batch, that is a parallel computation described in Section~\ref{ImplemDet}.

In addition, complying with the norm-constrained set-up~\cite{wang2017normface,ranjan2017l2}, both normal vectors and embeddings are kept in their unit form, meaning that the calculated projections correspond to the cosine values.
This fact, restricts the projections to $[-1,1]$ with the lower and upper bounds, indicating complete opposition and overlap, respectively.
Empirically, this case of complete features alignment in space appears infeasible, and the original optimization objective of SVM needs relaxation.
Therefore, we introduce a margin parameter $m$ to the SVM's cost function described in Section~\ref{ImplemDet}.

\subsection{Notation}

For the purposes of clarification, the imlementation description in Section~\ref{ImplemDet} adopts the Einstein summation convention~\cite{einstein1916foundation}. To that end, the double appearance of an index variable in a single term of an equation stands for the summation of the corresponding term over all the values of that index. Hence, for $\alpha,\beta$ of length $n$ we write:
\begin{equation}
\sum_{i=1}^{n}{\alpha_i\beta^i} \equiv \alpha_i\beta^i.\label{def}
\end{equation}
Note that the superscripts imply contra-variant (column) \textit{vectors}, while the subscripts indicate co-variant (row) vectors, \textit{aka. covectors}, thus having:
\begin{equation}
\alpha_i\beta^i = [\alpha_{1} \hdots \alpha_{n}]
  \begin{bmatrix}
    \beta^{1} \\ \vdots \\ \beta^{n}
  \end{bmatrix}.\label{vec-covec}
\end{equation}
\textit{Exception:} The \textit{dot symbol} ($\cdot$) is employed before an index that appears twice in a single term to indicate element-wise multiplication between the values of that index. Hence, the element-wise product of two vectors $\alpha, \beta$ of size $n$ is written:
\begin{equation}
\alpha^{\cdot i}\beta^{\cdot i} = [\alpha^1\beta^1 \hdots \alpha^n\beta^n].
\end{equation}

\subsection{Implementation Details}\label{ImplemDet}
\textit{HASeparator} can be easily implemented as a custom loss function in various deep learning frameworks, like Tensorflow~\cite{abadi2016tensorflow} and Pytorch~\cite{paszke2017automatic}.
In order to support GPU-enabled operations, its formulation adopts the entire tensor representations of the required variables instead of exploiting indexed terms.
A description of the layer inputs is illustrated in Table~\ref{tab1}.
Note that batch dimension $B$ is included in the following analysis, as well.
For each dimension, we introduce a specific index, $i$ for the batch dimension $B$, $k$ for the neurons dimension $N$, and $j$ for the classes one $C$.

\begin{table}[]
\caption{\textit{HASeparator} Inputs}
\begin{center}
\begin{tabular}{c|c|c}
\textbf{Variable} & \textbf{\textit{Symbol}}& \textbf{\textit{Shape}}$^{*}$\\
\hline
Feature Embeddings & $E$ & $B\times N$ \\
Layer Weights & $W$ & $N \times C$ \\
Target Labels & $L$ & $B\times 1$ \\
Feature Scaler & $\sigma$ & $1$ \\
Margin Parameter & $\mu$ & $1$ \\
\multicolumn{3}{l}{$^{*}$ $B$: batch size, $N$: layer units, $C$: number of classes.}
\end{tabular}
\label{tab1}
\end{center}
\end{table}

To begin with, an $l_2$-normalization over the neurons axis $N$ is performed both to the feature embeddings $E$ and the layer weights $W$~\cite{wang2017normface,wang2018cosface,deng2019arcface}, producing their corresponding normalized representations $\widehat{E}$ and $\widehat{W}$.
Following~\cite{deng2019arcface}, the matrix multiplication of the above tensors estimates the cosines between the embeddings and the weights.
Then, a scalar multiplication with the feature scaler $\sigma$ is performed to calculate the layer's output logits $O$ given a specific batch.
The above are summarized as follows:
\begin{equation}
O^{i}_{\ j} = \sigma\ \widehat{E}^{i}_{\ k} \widehat{W}^{k}_{\ j}.\label{logits}
\end{equation}
Along with the output logits, \textit{HASeparator} aims at calculating the total projections $P=\{p^i_{\ j}, \forall i=1,..,B\ \text{and}\ j=1,...,C\}$ of the normalized embeddings on the normal vectors of the separation hyperplanes of their target class.
To achieve that, we gather the target class's weight vector $\hat{w}^k_{\ L^i}$ for each sample $i$ based on $L$, and we store them in a tensor $T$ of size \mbox{$B\times N\times 1$}, where $1$ is kept to indicate the classes dimension.
Since the tensors $\widehat{W}$ and $T$ need to be subtracted for the calculation of the normal vectors, they have to be reformed, in order to display compatible shapes.
The above procedure is widely known as \textit{broadcasting} in deep learning nomenclature.
For that purpose, the normalized weight tensor $\widehat{W}$ is replicated by $B$ times and appended, forming the expanded tensor $\widehat{W}^{+}$ with shape \mbox{$B\times N\times C$}.
A similar procedure is applied in the classes dimension to replicate $T$ to $T^{+}$ with shape \mbox{$B\times N\times C$}.
Note that the above two operations are not computationally intensive since broadcasting is usually performed by default in most of the available deep learning frameworks.
Finally, we can calculate the separation hyperplanes, by applying:
\begin{equation}
H = \widehat{W}^{+}-T^{+}.
\end{equation}
The size of \mbox{$B\times N\times C$} that $H$ displays corresponds to the per-sample $C$ normal column vectors $h^{i}_{\ j}\in \mathbb{R}^N$.
In order to calculate the projection $P$, we calculate the tensor with the normal vectors $\widehat{H}$ through $l_2$-normalization over the $N$ axis.
Then, $P$ arises from the operation below:
\begin{equation}
P^{i}_{\ j} = \widehat{E}^{\cdot i}_{\ \ k} \widehat{H}^{\cdot ik}_{\ \ \ j},\label{einsum}
\end{equation}
where $\cdot i$ indicates element-wise multiplication over the batch axis $B$.
In~\eqref{einsum} we apply matrix multiplication between each embedding from the batch and the $C$ hyperplanes of the target class, producing $C$ projections for each embedding. 
Hence, the size of $P$ is $B\times C$.
Equation~\eqref{einsum} can be quickly performed through the \texttt{einsum} method of common platforms~\cite{abadi2016tensorflow,paszke2017automatic}.

Consequently, we modified the cost function of SVM introducing a relaxation margin parameter $m$ as mentioned in Section~\ref{Intuit}. The proposed function $J$ is element-wise applied to the projection tensor $P$, thus calculating the costs of each projection. Its equation is:
\begin{equation}
J(p^i_{\ j}) = m - \min{(p^i_{\ j},m)},\label{costeq}
\end{equation}
with $m\in (0,1]$. 
The larger the value of $m$, the more rigorous the function $J$.
Then, the sum over the $C$ axis is calculated to define the loss of each embedding, and the total cost ($C_t$) arises from the mean value of the batch's embedding losses.
The above are performed through the following equation:
\begin{equation}
C_t = \frac{1}{B}1_iJ^i_{\ j}1^j,\ \text{with } 1_i = [1 \hdots 1]\ \text{and } 1^j = [1 \hdots 1]^{T}
\end{equation}
constituting the vector and covector of length $B$ and $C$, respectively.
Eventually, $C_t$ is added to the regular crossentropy loss $C_c$~\cite{kansizoglou2019active}, forming the total loss $C_{all} = C_t+C_c$.

\section{Experiments}\label{Exper}

\begin{figure}[]
    \centering
    \begin{subfigure}[b]{0.24\textwidth}
        \centering
        \includegraphics[width=\textwidth]{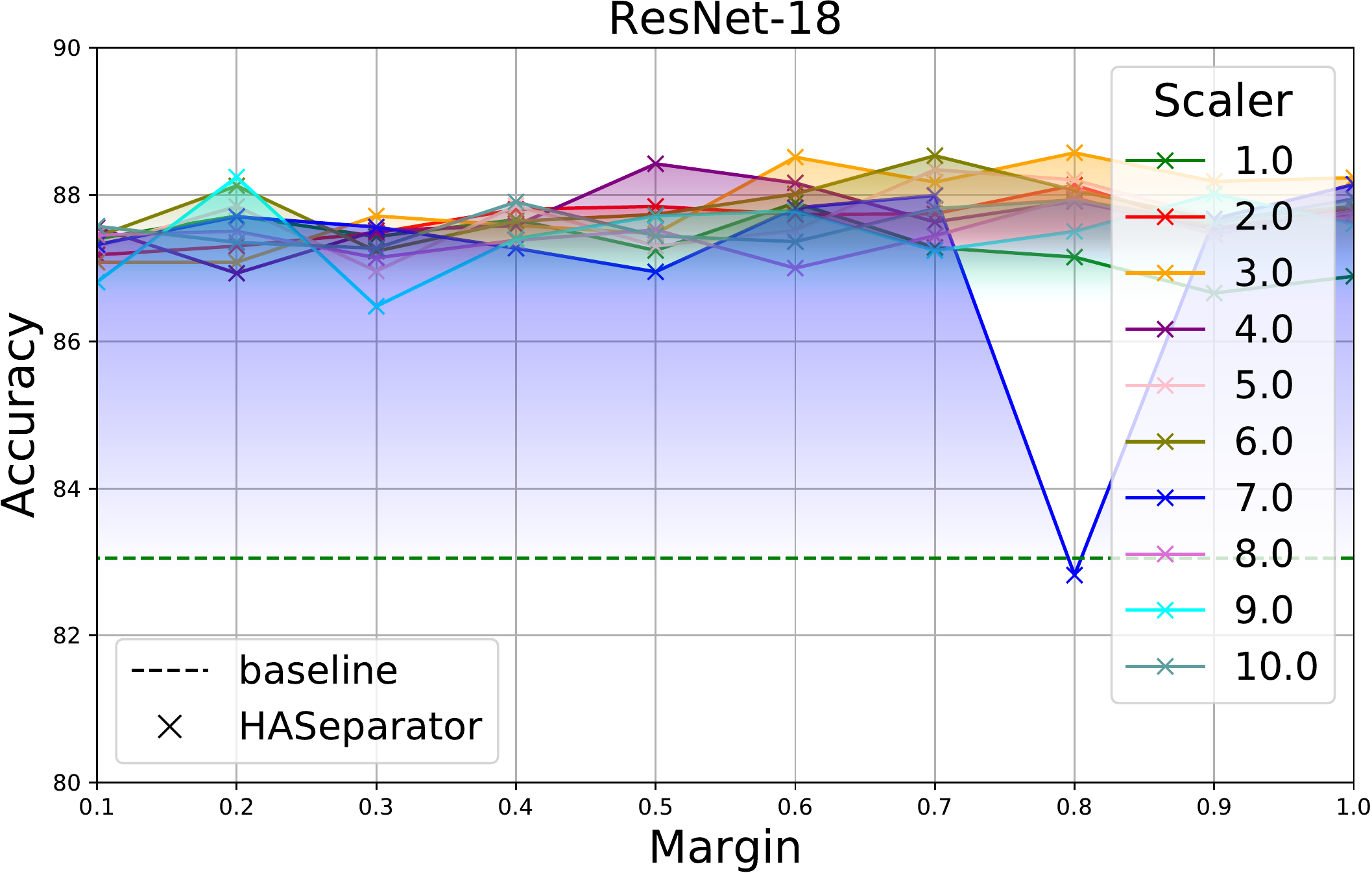}          
    	\caption[]{ResNet-18 on C10}
    	\label{fig:subAcca}
    \end{subfigure}
    \centering
    \begin{subfigure}[b]{0.24\textwidth}
		\centering    	
        \includegraphics[width=\textwidth]{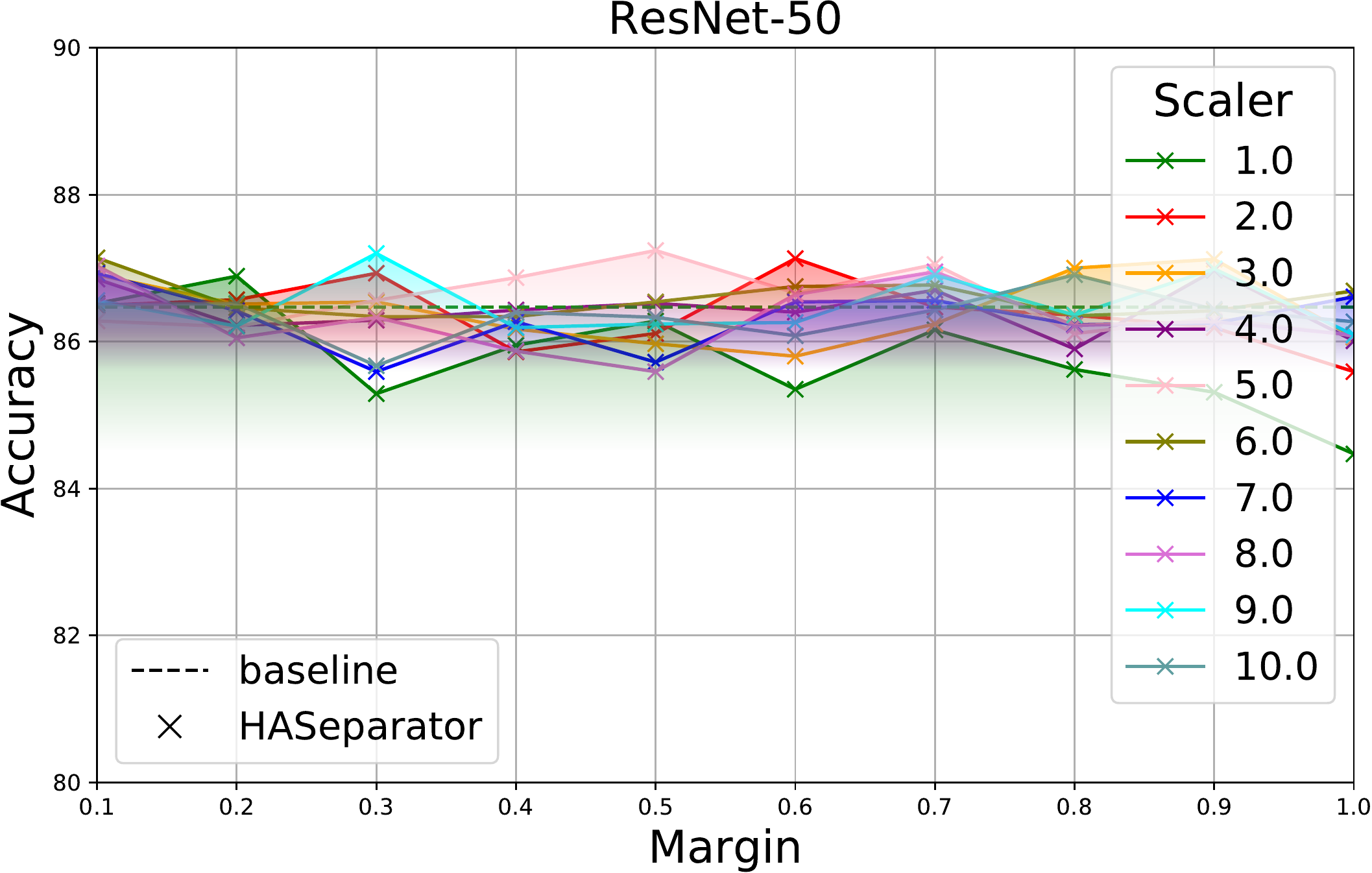}          
    	\caption[]{ResNet-50 on C10}
    	\label{fig:subAccb}
    \end{subfigure}
    \centering
    \vskip\baselineskip
    \centering
    \begin{subfigure}[b]{0.24\textwidth}
		\centering    	
        \includegraphics[width=\textwidth]{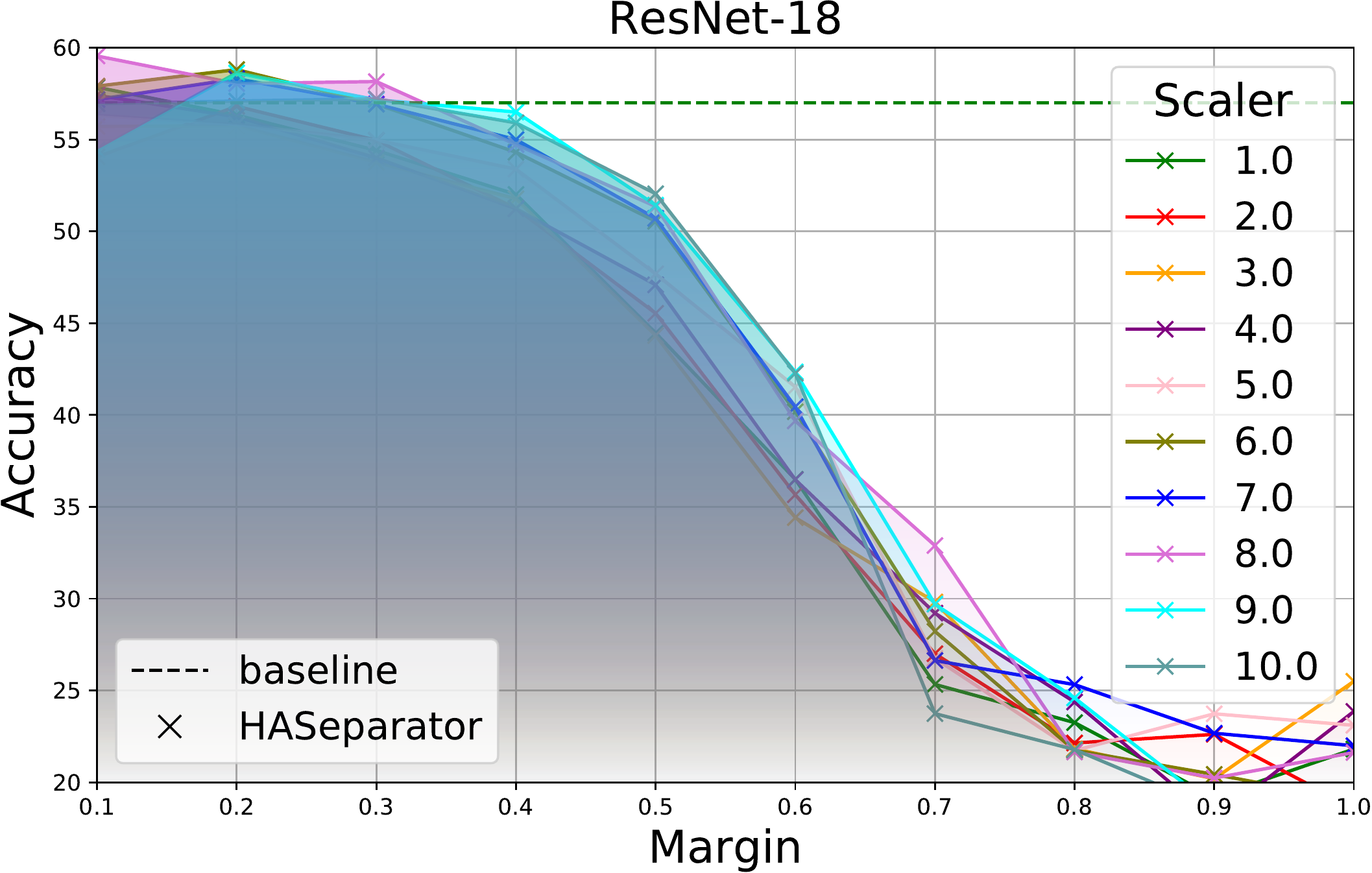}          
    	\caption[]{ResNet-18 on C100}
    	\label{fig:subAccc}
    \end{subfigure}
    \centering
     \begin{subfigure}[b]{0.24\textwidth}
		\centering    	
        \includegraphics[width=\textwidth]{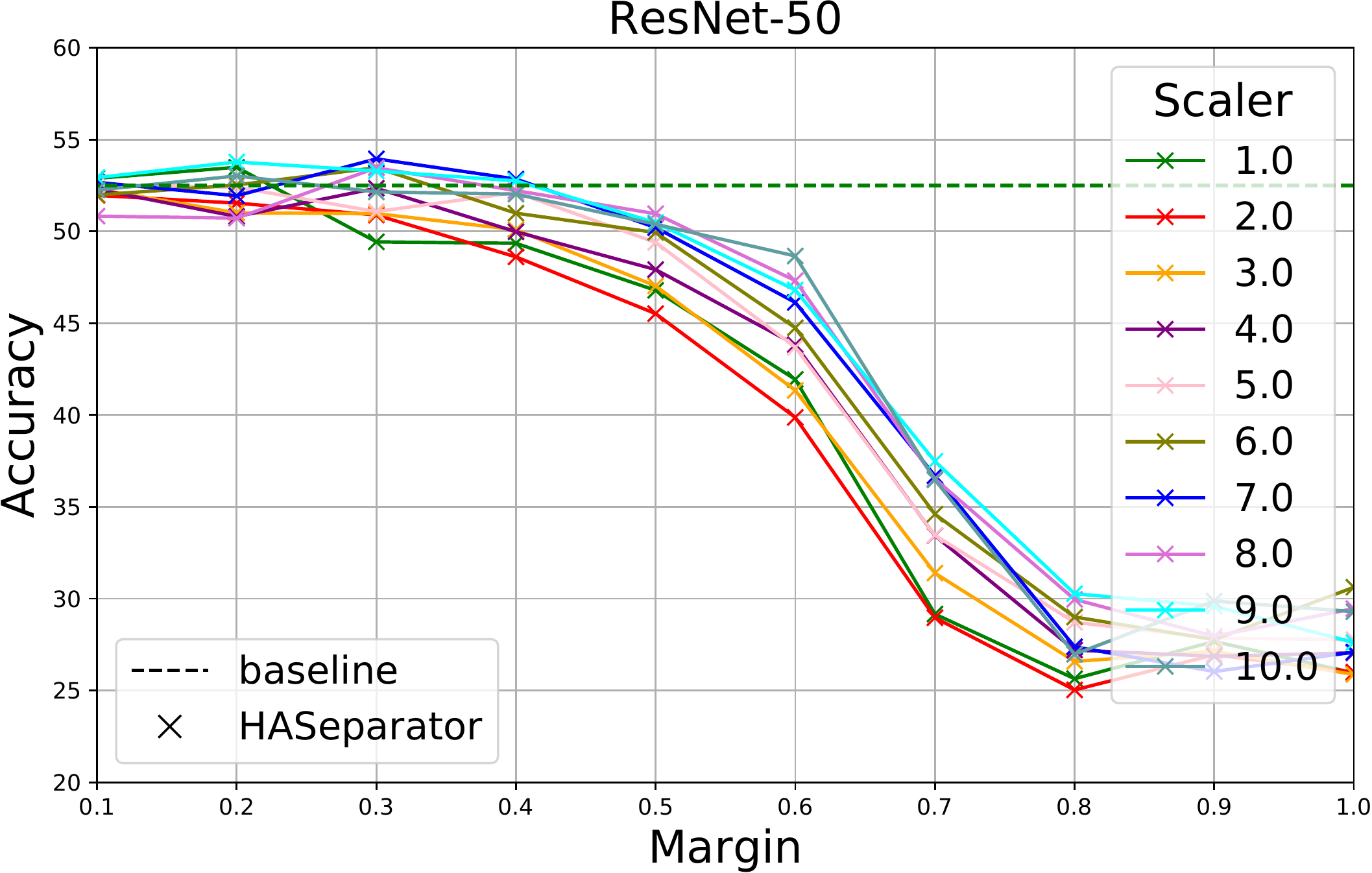}          
    	\caption[]{ResNet-50 on C100}
    	\label{fig:subAccd}
    \end{subfigure}
    \caption[]
    {\small Accuracy results for different values of scaler $\sigma$ and margin $m$.}
    \label{fig:Acc}
\end{figure}

We experimentally demonstrate the efficiency of \textit{HASeparator} in feature discrimination on designated benchmarks, exploiting two variants of the ResNet architectures~\cite{he2016deep}, \textit{viz.} ResNet-18 and ResNet-50.
Note that in the case of ResNet-50, we employ an additional dense layer with $64$ units before the classification one to reduce the feature space dimension $N$. In order to place our contribution within the state-of-the-art, we compare our method against  the broadly established \textit{ArcFace}~\cite{deng2019arcface} solution, proving the produced features' discrimination capabilities, while still retaining the same accuracy results during classification.

\begin{figure}[]
    \centering
    \begin{subfigure}[b]{0.235\textwidth}
        \centering
        \includegraphics[width=\textwidth]{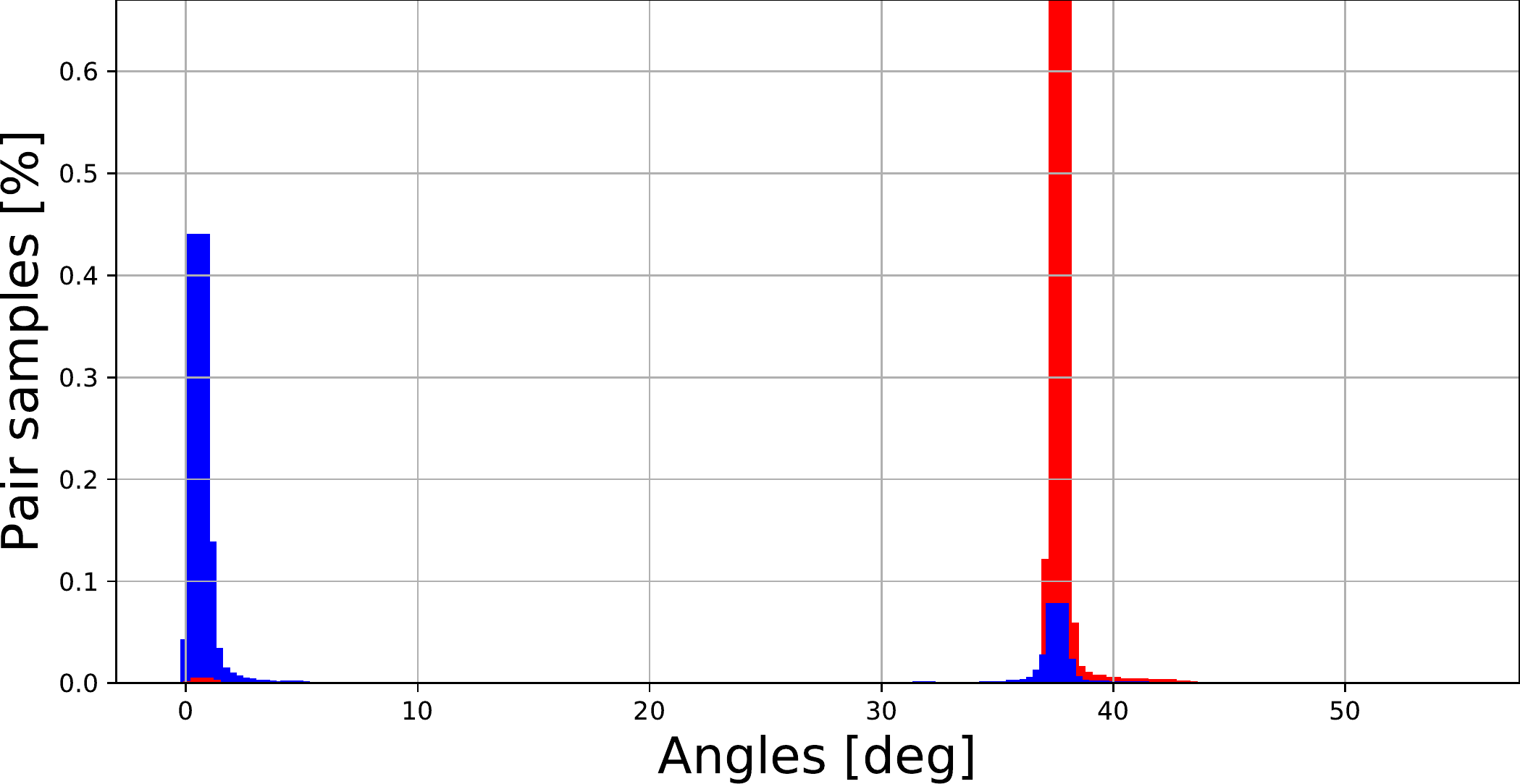}          
    	\label{fig:subHista}
    \end{subfigure}
    \centering
    \begin{subfigure}[b]{0.235\textwidth}
		\centering    	
        \includegraphics[width=\textwidth]{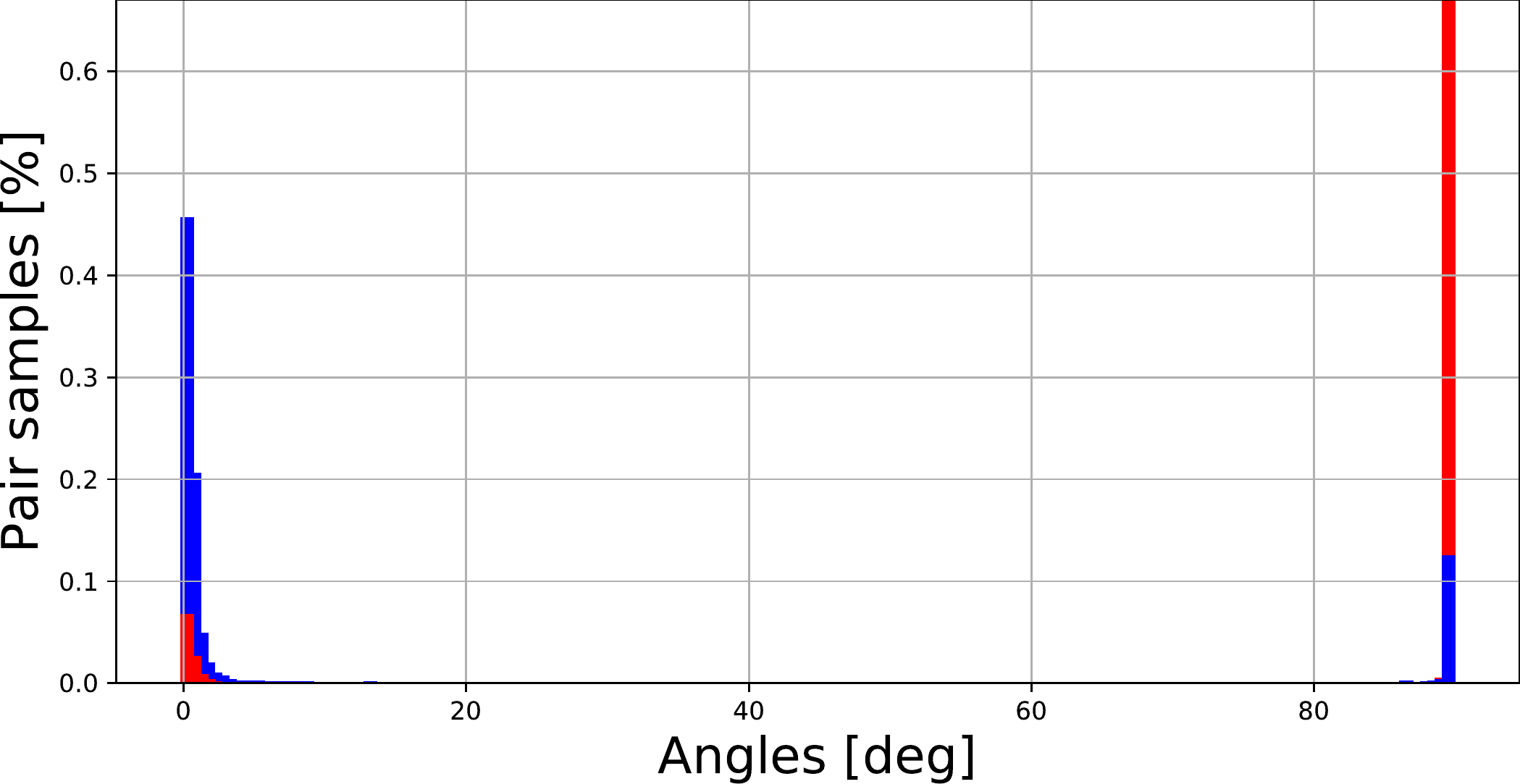}          
    	\label{fig:subHistb}
    \end{subfigure}
    \caption[]{\small Histograms regarding the angular distributions of positive and negative pairs of feature embeddings on C10.}
    \label{fig:Hist}
\end{figure}

\begin{figure*}[h]
    \centering
    \begin{subfigure}[b]{0.49\textwidth}
		\centering    	
    	\begin{subfigure}[b]{0.493\textwidth}
        	\centering
        	\includegraphics[width=\textwidth]{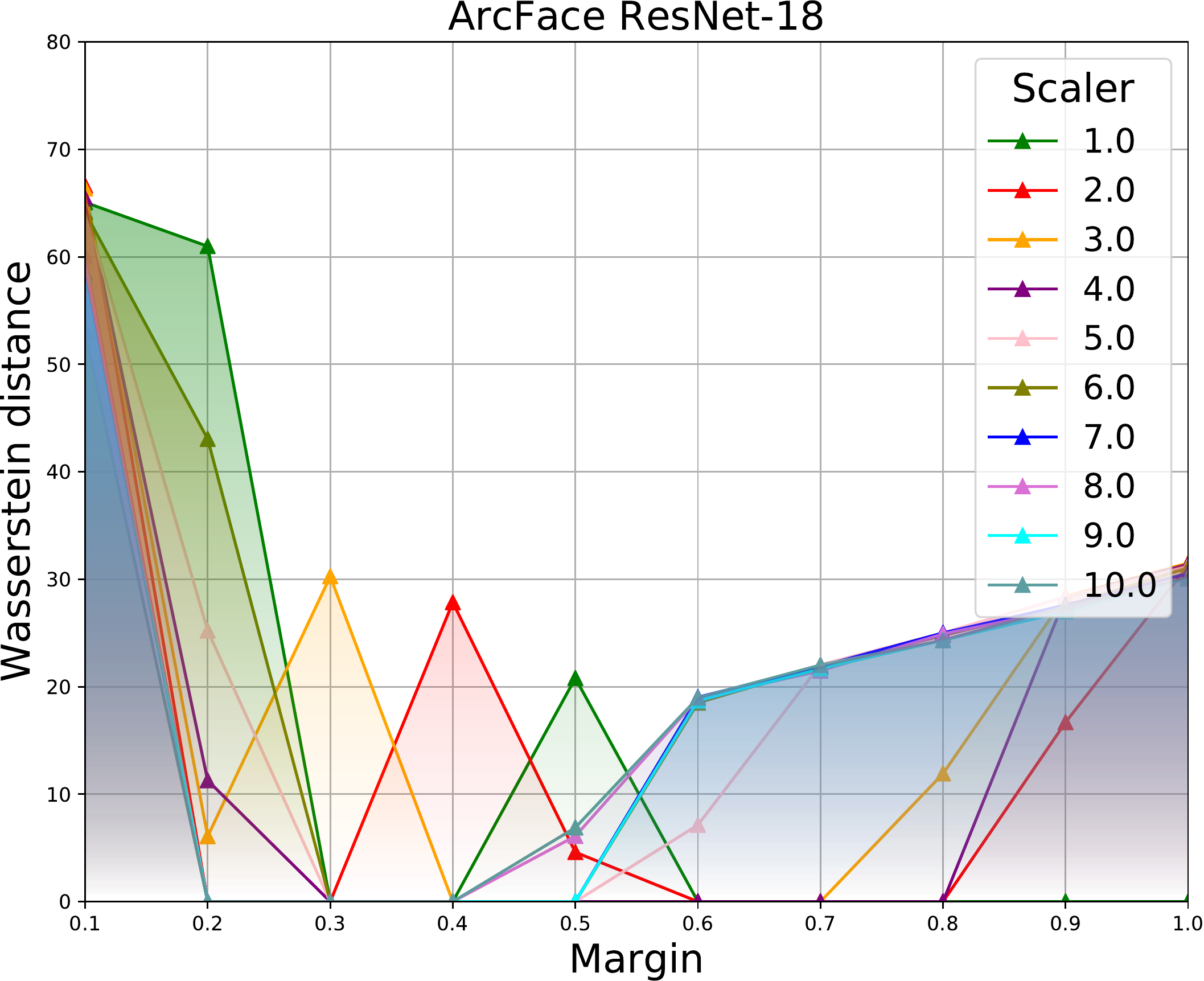}       
    	\end{subfigure}
    	\begin{subfigure}[b]{0.493\textwidth}
        	\centering
        	\includegraphics[width=\textwidth]{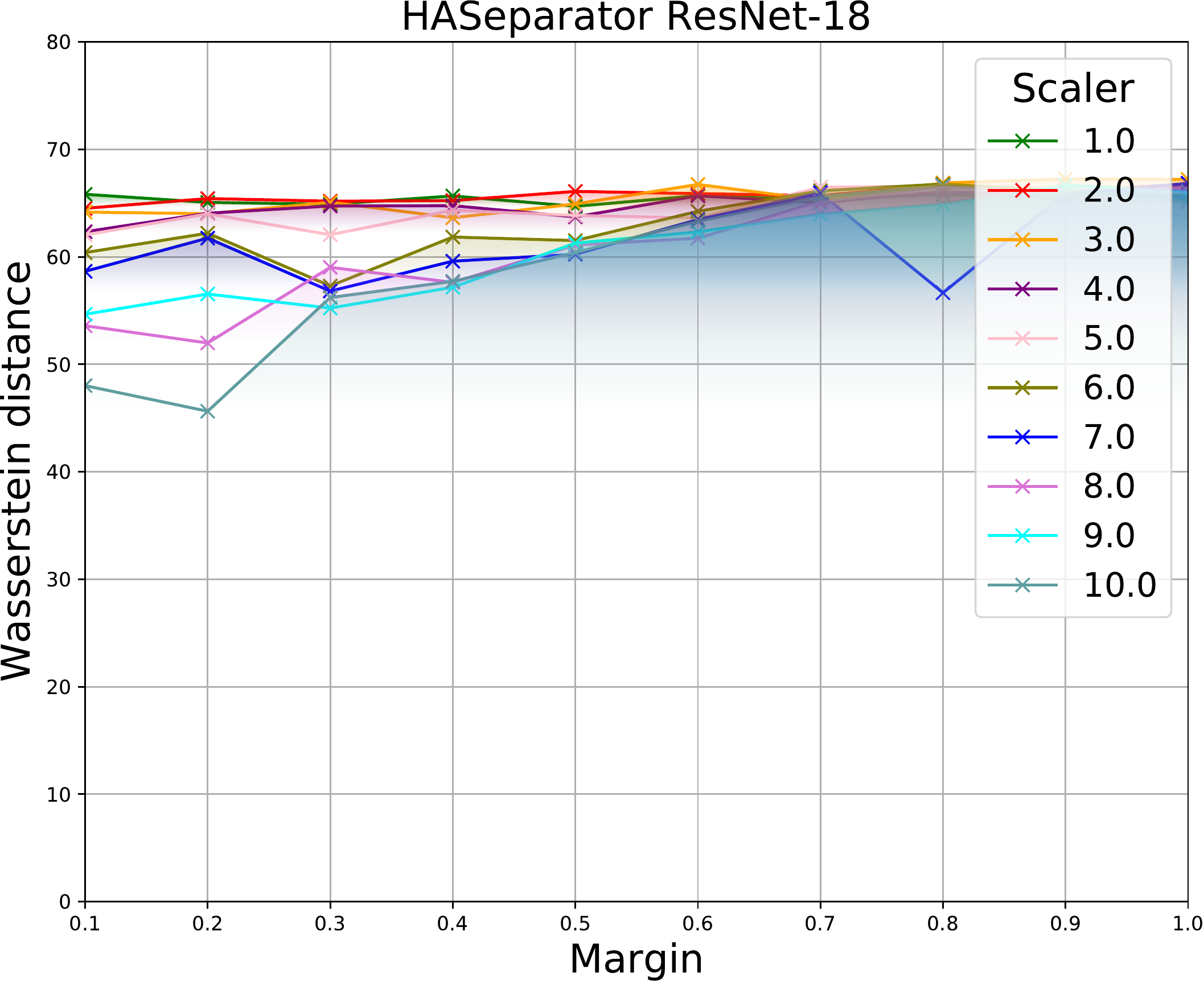}          
    	\end{subfigure}
    	\caption[]{ResNet-18 on C10}
    	\label{fig:sub3a}
    \end{subfigure}
    \centering
    \begin{subfigure}[b]{0.49\textwidth}
		\centering    	
    	\begin{subfigure}[b]{0.493\textwidth}
        	\centering
        	\includegraphics[width=\textwidth]{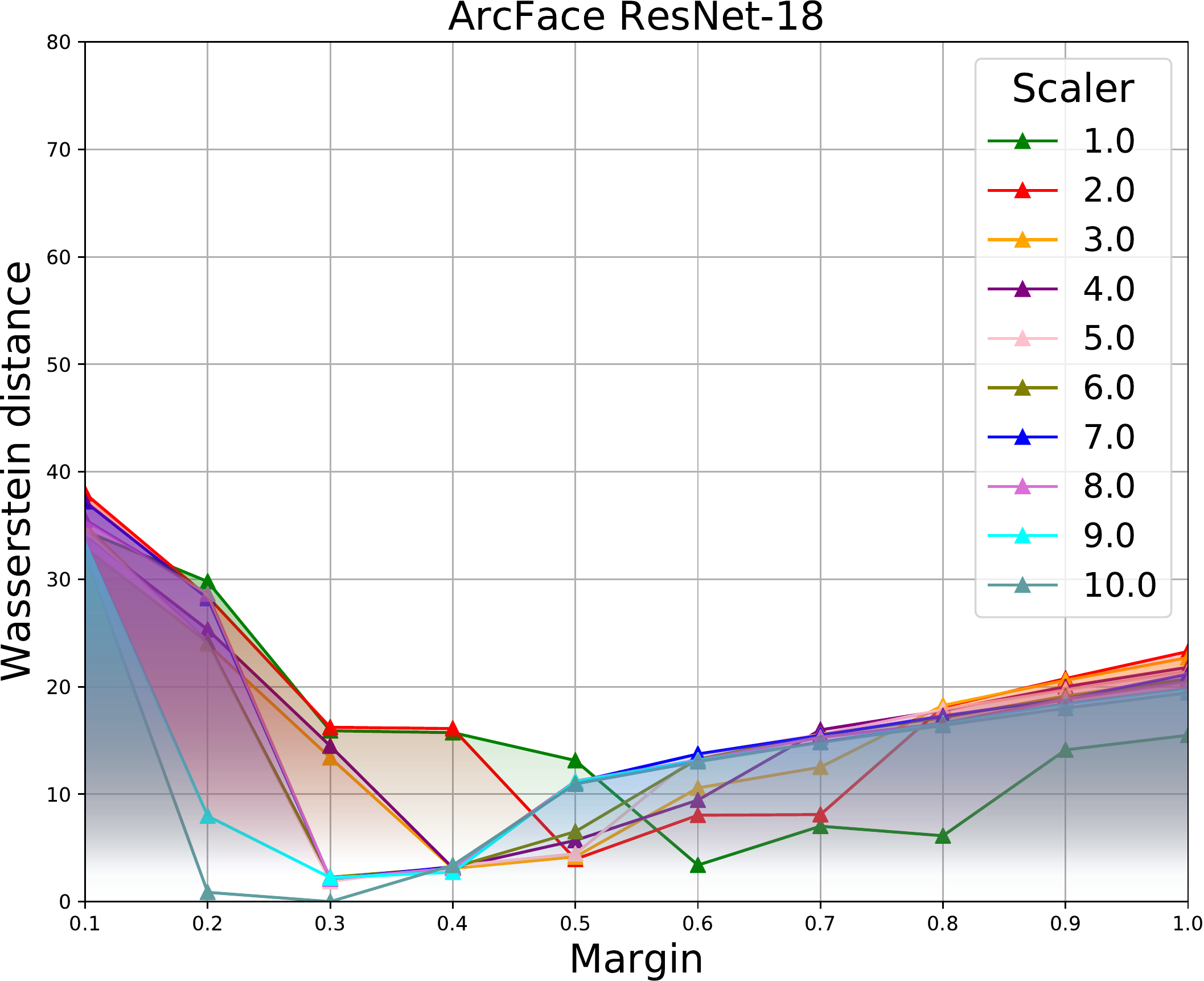}       
    	\end{subfigure}
    	\begin{subfigure}[b]{0.493\textwidth}
        	\centering
        	\includegraphics[width=\textwidth]{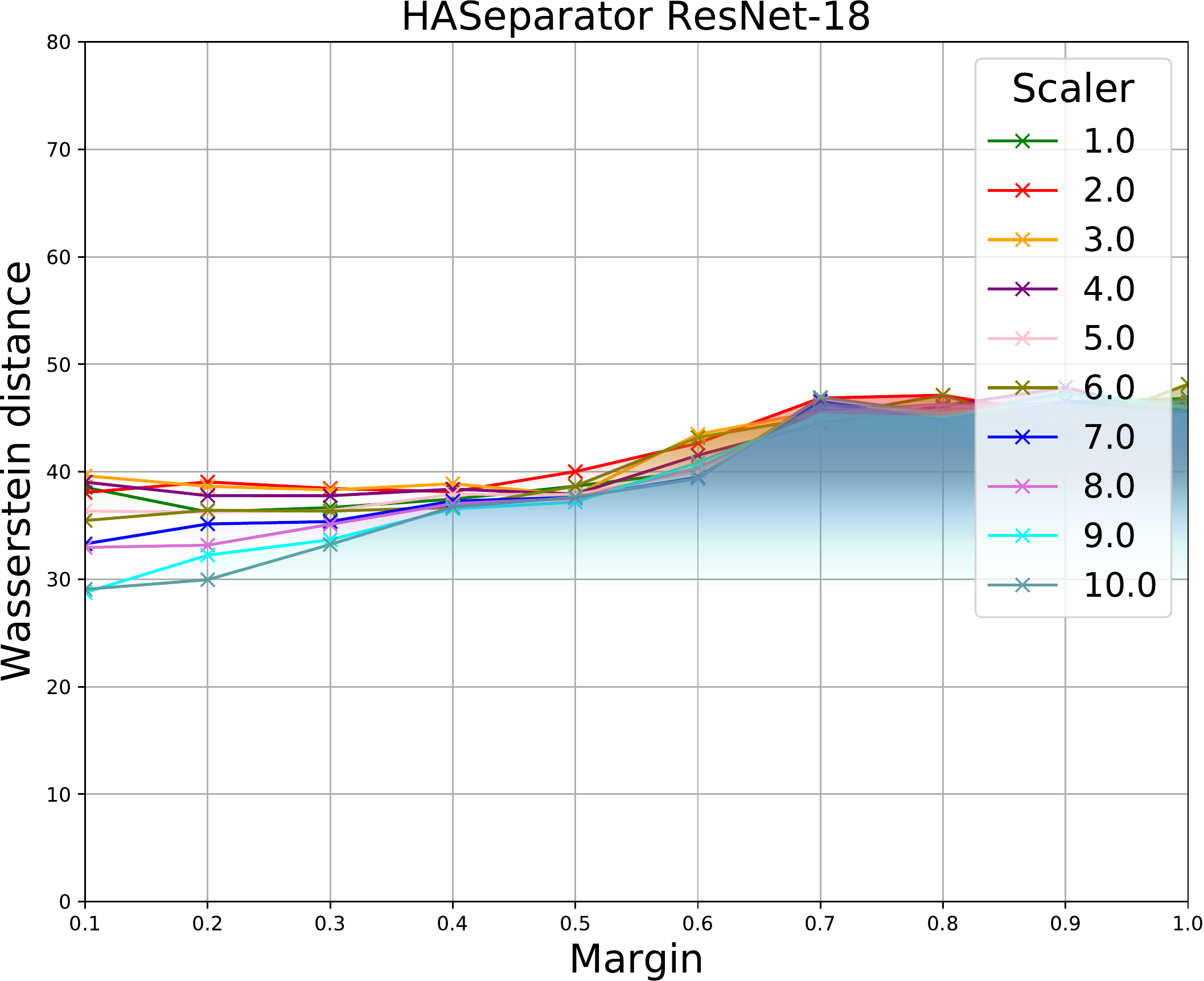}          
    	\end{subfigure}
    	\caption[]{ResNet-18 on C100}
    	\label{fig:sub3b}
    \end{subfigure}
    \centering
    \vskip\baselineskip
    \centering
    \begin{subfigure}[b]{0.49\textwidth}
		\centering    	
    	\begin{subfigure}[b]{0.493\textwidth}
        	\centering
        	\includegraphics[width=\textwidth]{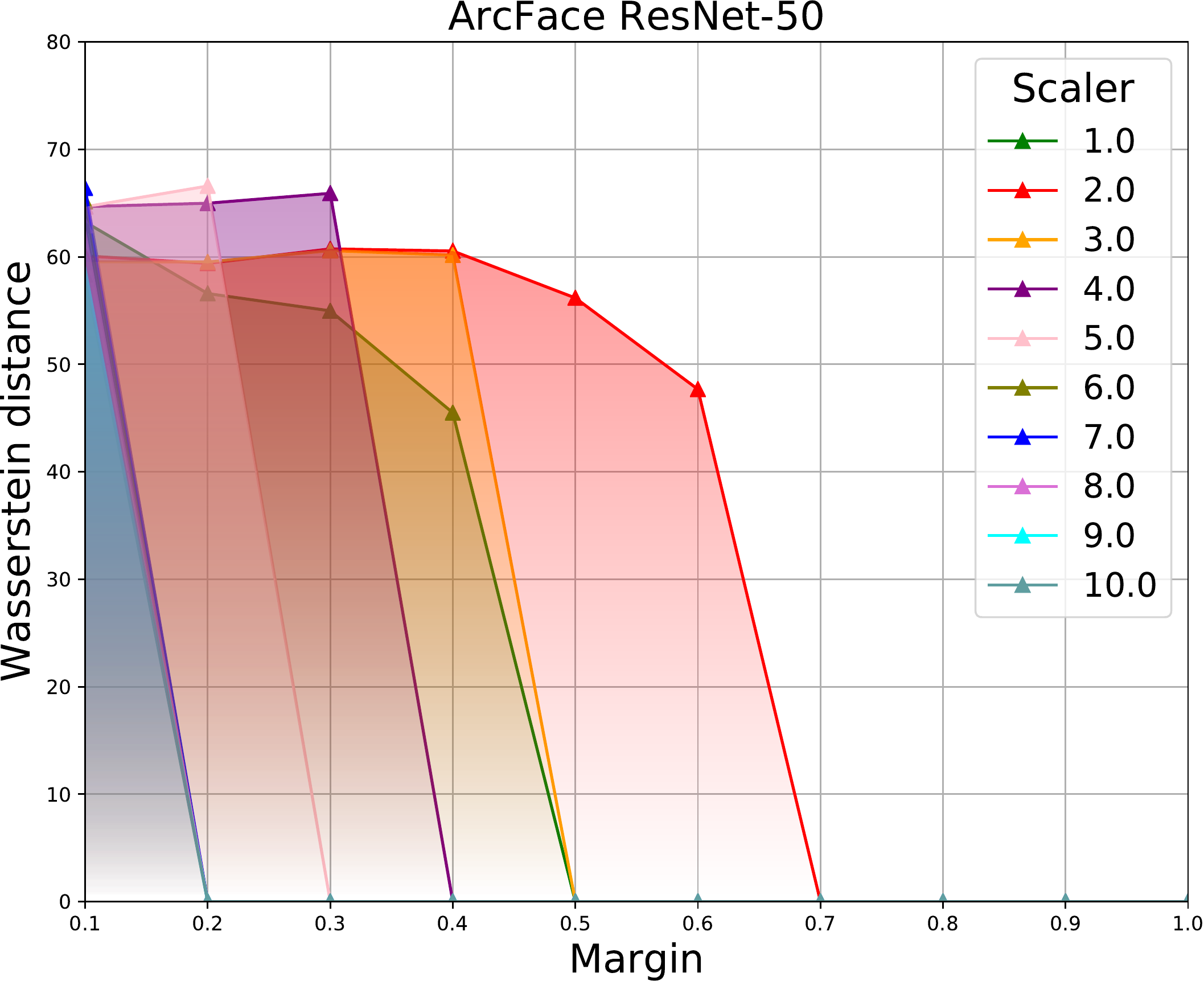}       
    	\end{subfigure}
    	\begin{subfigure}[b]{0.493\textwidth}
        	\centering
        	\includegraphics[width=\textwidth]{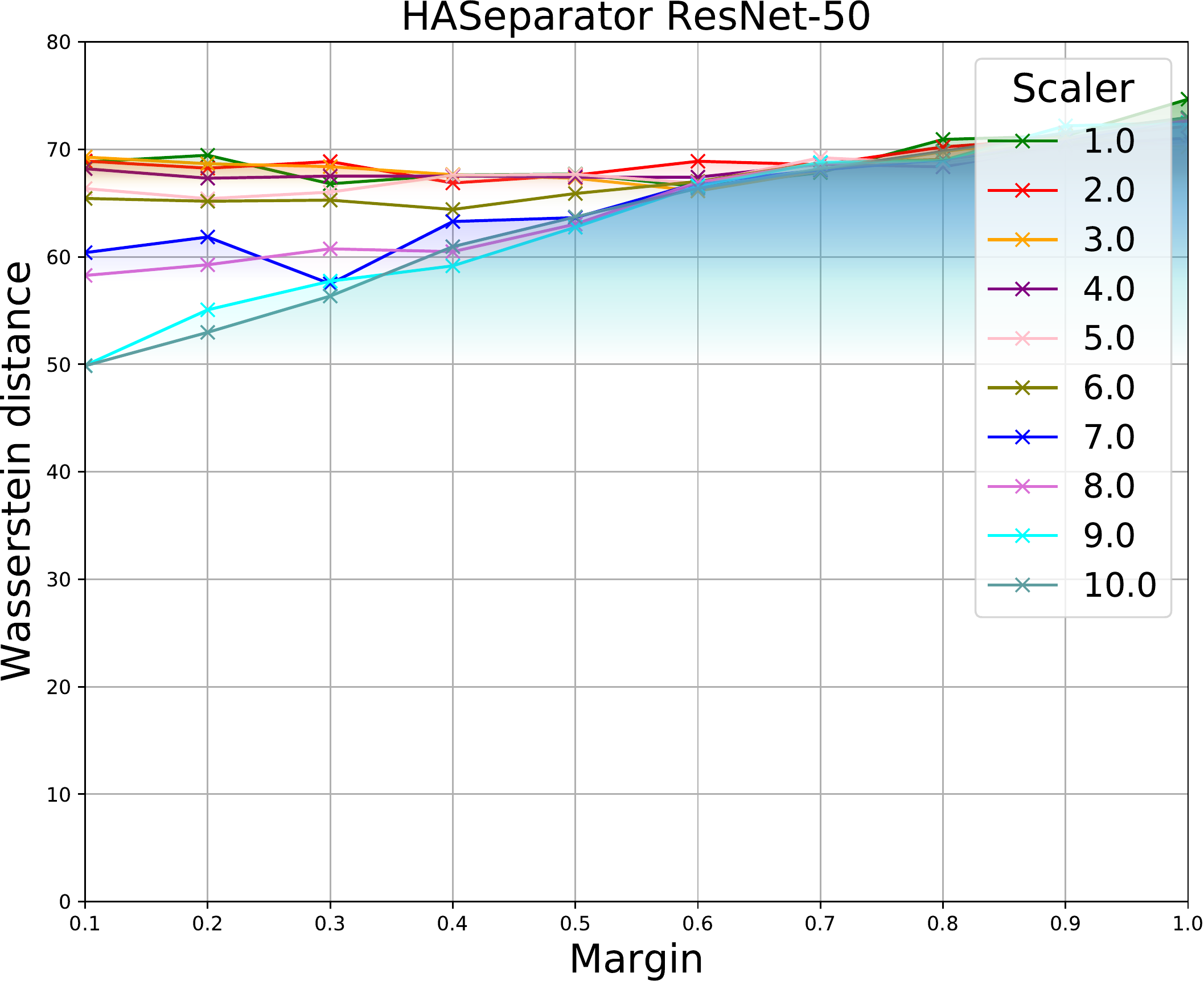}          
    	\end{subfigure}
    	\caption[]{ResNet-50 on C10}
    	\label{fig:sub3c}
    \end{subfigure}
    \centering
     \begin{subfigure}[b]{0.49\textwidth}
		\centering    	
    	\begin{subfigure}[b]{0.493\textwidth}
        	\centering
        	\includegraphics[width=\textwidth]{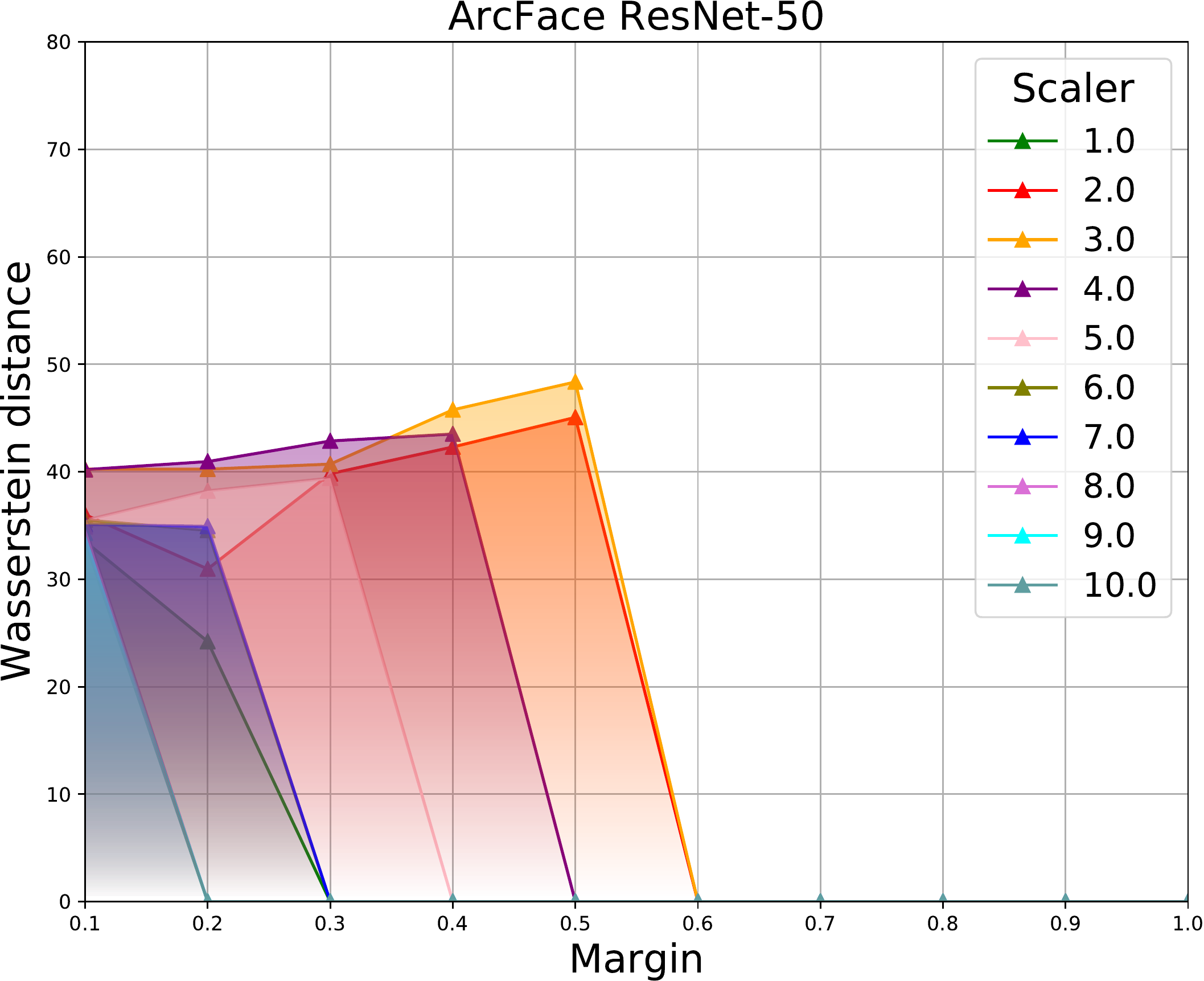}       
    	\end{subfigure}
    	\begin{subfigure}[b]{0.493\textwidth}
        	\centering
        	\includegraphics[width=\textwidth]{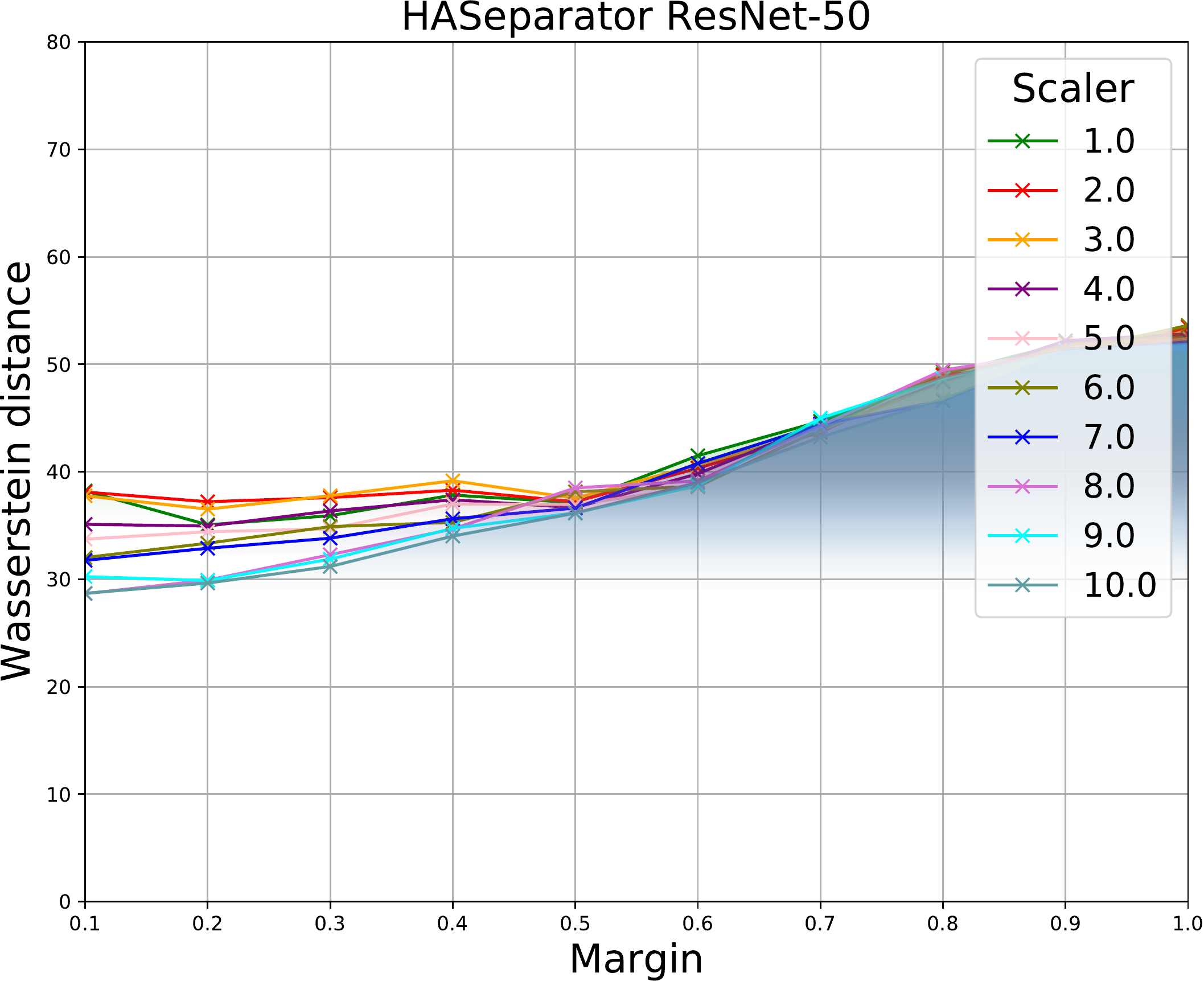}          
    	\end{subfigure}
    	\caption[]{ResNet-50 on C100}
    	\label{fig:sub3d}
    \end{subfigure}
    \caption[]
    {\small The obtained Wasserstein distances ($D_{EM}$) between the histograms of positive and negative pairs' angular distributions. Graphs compare the outputs of \textit{ArcFace} and the proposed \textit{HASeparator} for different values of scaler $\sigma$ and margin parameter $m$ on C10 and C100, using ResNet-18 (\textit{first row}) and ResNet-50 (\textit{second row}).}
    \label{fig:3}
\end{figure*}

\subsection{Datasets and Training Setups}\label{DataSet}

\subsubsection{CIFAR~\cite{krizhevsky2009learning}}
consists of two individual datasets; \mbox{CIFAR-10} (C10) with 10 target identities and \mbox{CIFAR-100} (C100) with 100.
Both of them comprise 50,000 training and 10,000 testing RGB images of $32\times 32$ pixels.
For preprocessing, each image is linearly scaled to display a mean value of $0$ and a variance of $1$.
During training on C10 and C100, we follow the configuration proposed in~\cite{he2016deep}, employing a batch size of 128, a training duration of 64k iterations, and a stochastic gradient descend (SGD) optimizer with a weight decay of $10^{-4}$ and momentum of $0.9$.
Accordingly, the initial learning rate is set to $0.1$ and decreased by $10$ at 32k and 48k iterations.
No augmentation scheme has been adopted.

\subsubsection{Street View House Numbers (SVHN)~\cite{netzer2011reading}}
forms a colored image dataset of $10$ target classes. 
It consists of 73,257 training and 26,032 testing images of size $32\times 32$ pixels.
Note that we do not exploit the additional 531,131 images of the dataset, and no data augmentation is employed.
For preprocessing, we apply the same standardization rule with C10 and C100.
During training, we also adopt the same SGD optimizer and batch size with CIFAR.
Finally, the training procedure lasts 40 epochs with an initial learning rate of 0.1 that decreases by $10$ at epochs $20$ and $30$~\cite{huang2017densely}.

\subsection{Classification Accuracy on CIFAR}

The primary goal of \textit{HASeparator} is to ensure at least the same classification performance as the simple softmax loss.
Hence, during our experimentation, accuracy constitutes the first evaluation metric defining the acceptability of the produced results.
In Fig.~\ref{fig:Acc}, we present the obtained accuracy curves for a wide range of scaler $\sigma$ and margin parameter $m$ variations.
The above procedure was performed on C10 and C100, both for ResNet-18 and ResNet-50.
Regarding C10, we observed quite similar classification performances with small fluctuations under $\sigma$ and $m$ changes, with most of them being equal or even higher from the baseline softmax loss performance.
However, in the case of C100, we noticed a gradually classification accuracy reduction with the increase of the margin parameter $m$, which indicates that for higher number of target classes the relaxation of $J$ is essential, in order to sustain the classification performance.

\subsection{Evaluation Metrics for Feature Discrimination}\label{metrics}

\begin{figure*}[h]
    \centering
    \begin{subfigure}[b]{0.48\textwidth}
        \centering
        \includegraphics[width=0.63\textwidth]{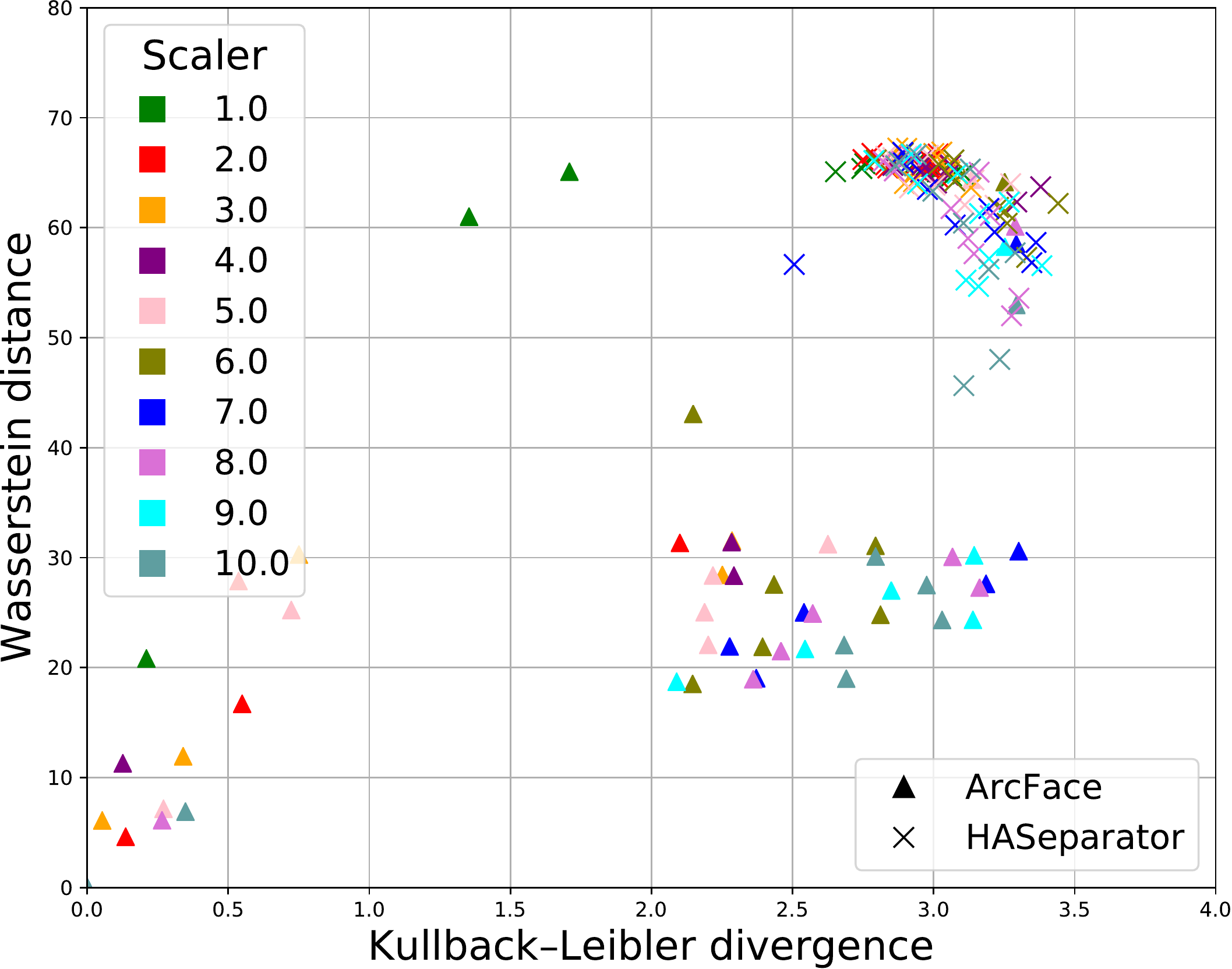}          
    	\caption[]{ResNet-18 on C10}
    	\label{fig:sub4a}
    \end{subfigure}
    \centering
    \begin{subfigure}[b]{0.48\textwidth}
		\centering    	
        \includegraphics[width=0.63\textwidth]{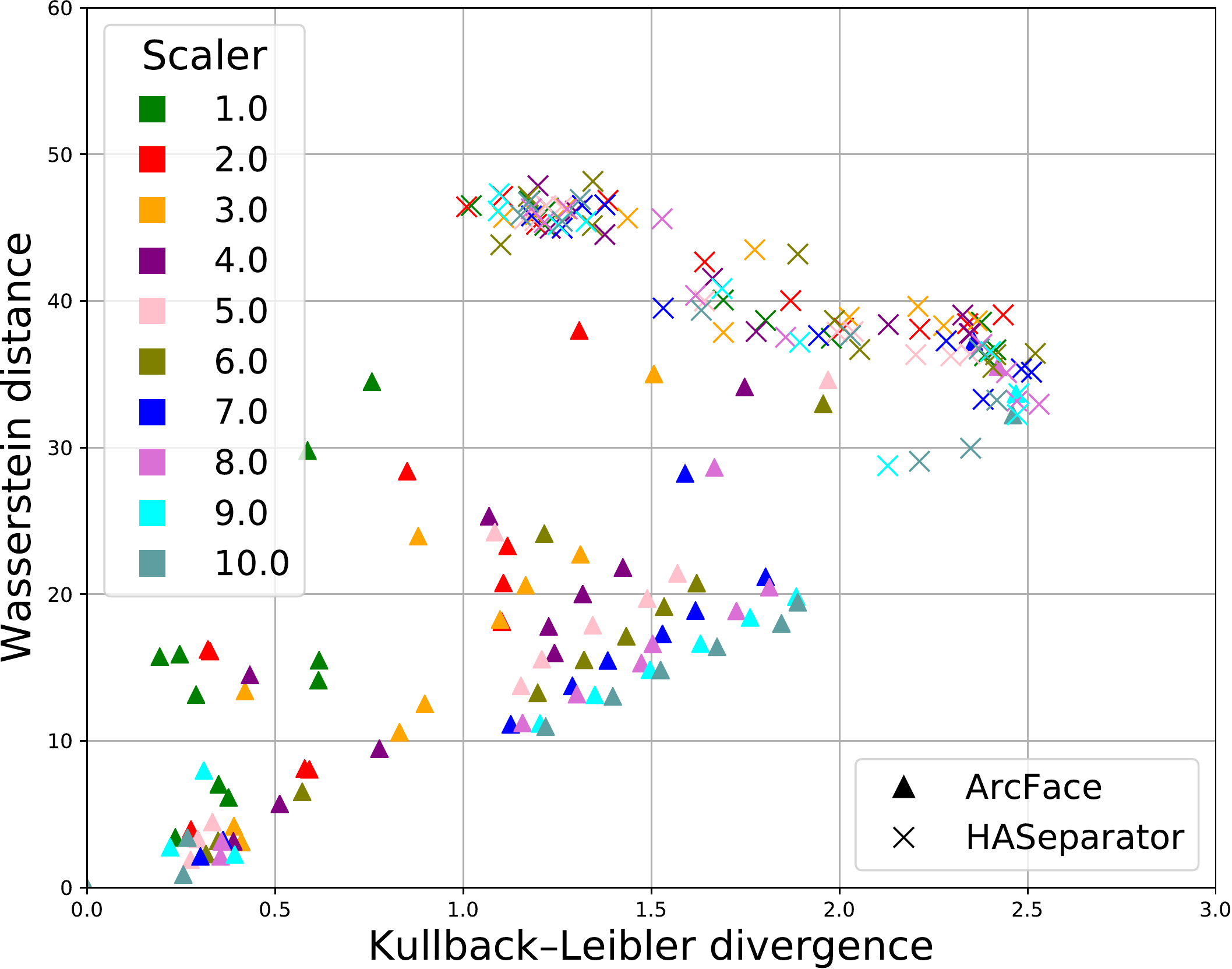}          
    	\caption[]{ResNet-18 on C100}
    	\label{fig:sub4b}
    \end{subfigure}
    \centering
    \vskip\baselineskip
    \centering
    \begin{subfigure}[b]{0.48\textwidth}
		\centering    	
        \includegraphics[width=0.63\textwidth]{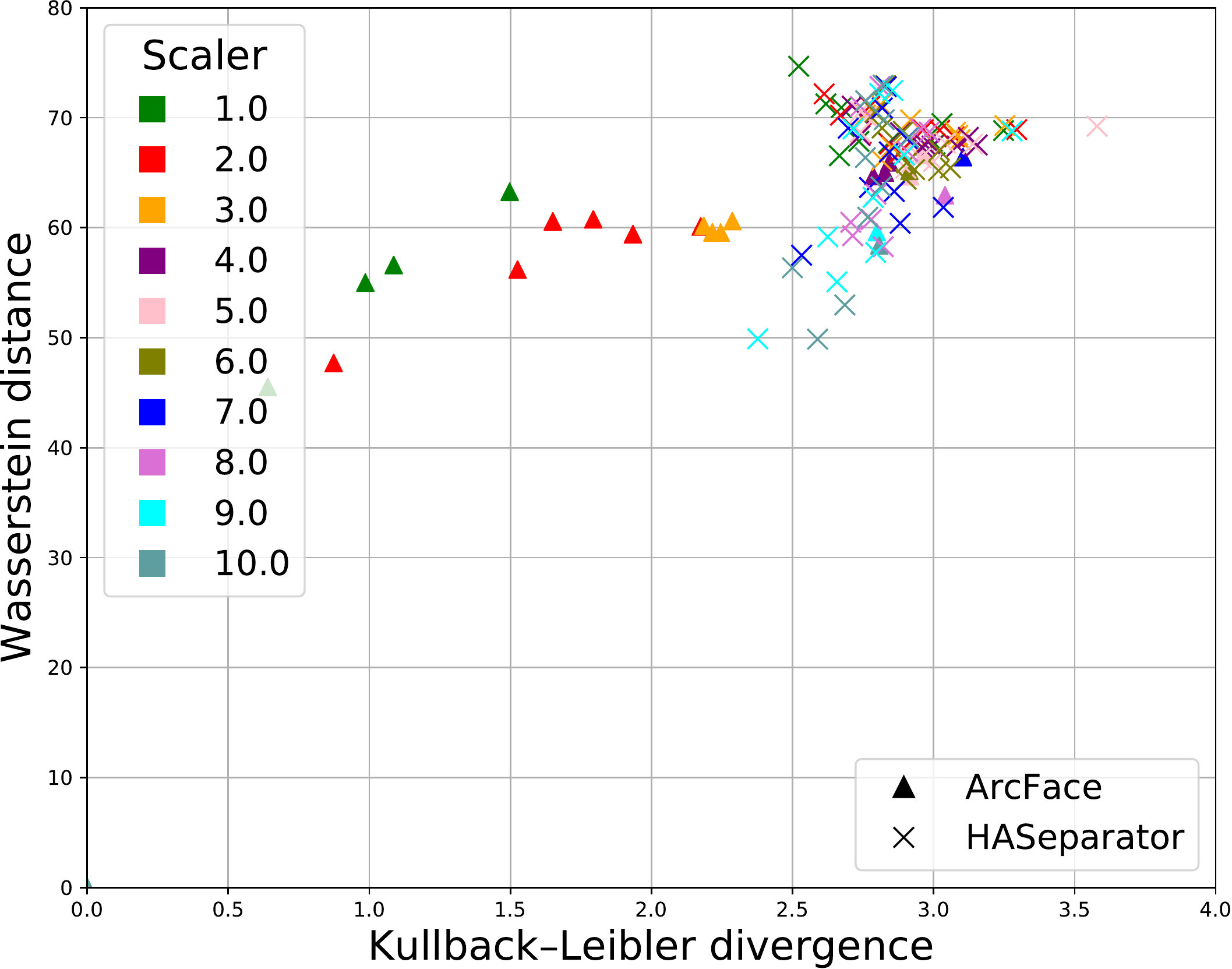}          
    	\caption[]{ResNet-50 on C10}
    	\label{fig:sub4c}
    \end{subfigure}
    \centering
     \begin{subfigure}[b]{0.48\textwidth}
		\centering    	
        \includegraphics[width=0.63\textwidth]{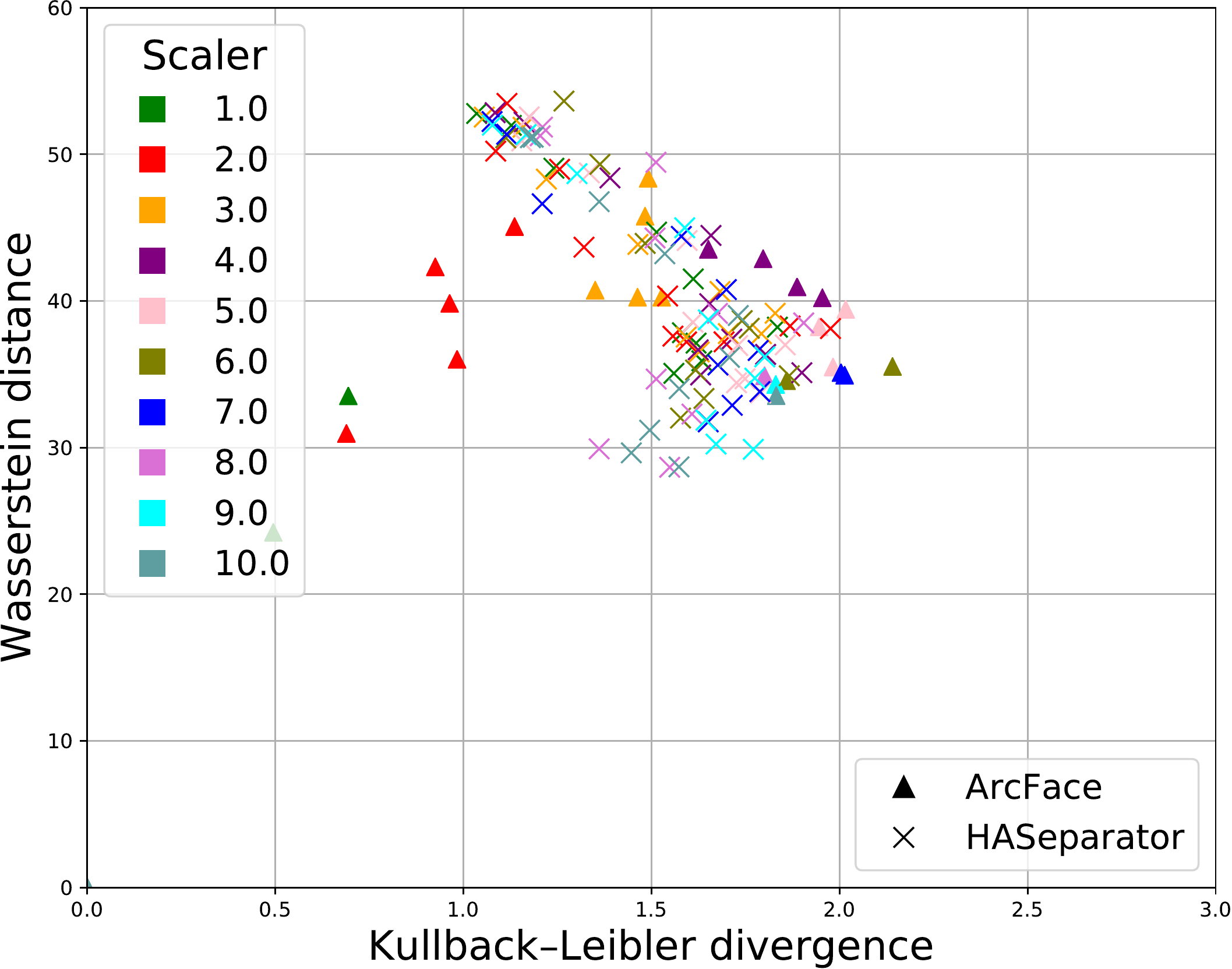}          
    	\caption[]{ResNet-50 on C100}
    	\label{fig:sub4d}
    \end{subfigure}
    \caption[]
    {\small Graphical assessment of different configurations and the general behavior of \textit{ArcFace} and \textit{HASeparator}. The horizontal axis corresponds to the obtained distributions' KL-divergence ($D_{KL}$), while the vertical one indicates their Wasserstein distance ($D_{EM}$).}
    \label{fig:4}
\end{figure*}

\begin{table*}
\centering
\caption{\small Top-2 configurations of \textit{ArcFace} and \textit{HASeparator} for ResNet-18 and ResNet-50, on C10 and C100.}
\label{table:Top2}
\resizebox{\linewidth}{!}{%
\renewcommand{\arraystretch}{1.2}
\begin{tabular}{cc|c|c|c|c|c|c|c|c|c|c|c|c|c|c|c}
\multirow{3}{*}{} & \multicolumn{8}{c|}{\textbf{C10}} & \multicolumn{8}{c}{\textbf{C100}} \\
 & \multicolumn{4}{c|}{\textbf{ResNet-18}} & \multicolumn{4}{c|}{\textbf{ResNet-50}} & \multicolumn{4}{c|}{\textbf{ResNet-18}} & \multicolumn{4}{c}{\textbf{ResNet-50}} \\
  \cline{2-17}
  & \multicolumn{2}{|c|}{\textit{ArcFace}} & \multicolumn{2}{c|}{\textit{HASeparator}} 
  & \multicolumn{2}{c|}{\textit{ArcFace}} & \multicolumn{2}{c|}{\textit{HASeparator}} 
  & \multicolumn{2}{c|}{\textit{ArcFace}} & \multicolumn{2}{c|}{\textit{HASeparator}}
  & \multicolumn{2}{c|}{\textit{ArcFace}} & \multicolumn{2}{c|}{\textit{HASeparator}}\\
\hline
\multicolumn{1}{c|}{$\sigma$} & $2.0$ & $4.0$ & $3.0$ & $3.0$ &
$5.0$ & $7.0$ & $10.0$ & $6.0$ &
 $7.0$ & $8.0$ & $1.0$ & $6.0$ &
 $7.0$ & $7.0$ & $1.0$ & $6.0$ \\
\multicolumn{1}{c|}{$m$} & $0.1$ & $0.1$ & $0.9$ & $1.0$ &
$0.2$ & $0.1$ & $1.0$ & $1.0$ &
 $0.1$ & $0.1$ & $1.0$ & $0.2$ &
 $0.1$ & $0.2$ & $0.2$ & $0.3$ \\
\hline
\multicolumn{1}{c|}{$D_{KL}$} & $2.879$ & $2.868$ & $2.873$ & $\textbf{2.905}$ &
$2.963$ & $\textbf{3.104}$ & $2.821$ & $2.830$ &
 $2.357$ & $2.421$ & $2.377$ & $\textbf{2.520}$ &
 $2.004$ & $\textbf{2.014}$ & $1.560$ & $1.866$ \\
\multicolumn{1}{c|}{$D_{EM}$} & $66.61$ & $66.34$ & $\textbf{67.24}$ & $67.21$ &
$66.59$ & $66.39$ & $\textbf{72.96}$ & $72.91$ &
 $37.22$ & $35.52$ & $\textbf{38.55}$ & $36.42$ &
 $\textbf{35.10}$ & $34.92$ & $35.06$ & $34.90$ \\
\end{tabular}}
\end{table*}

Aiming to assess the discrimination performance of \textit{HASeparator} we introduce the following evaluation metrics.

\subsubsection{Histograms}
In order to capture feature discrimination in a latent space, one can not rely on the classification accuracy since it measures the performance in a quantized network's output space.
Instead, knowing the angular property of the feature embeddings, we can calculate the angles between each embedding and the ones that belong to the same class (positive pairs), as well as the same embedding and the ones that belong to different classes (negative pairs).
In an optimal scenario, the histograms of those two angular distributions should be completely distinguishable and without any overlap, with the positive pairs' angular distribution concentrated at $0^{o}$, and the negative pairs' one at $90^{o}$.

\subsubsection{Kullback-Leibler divergence~\cite{kullback1951information}}
is introduced to quantify the divergence between the positive and negative pairs' angular distributions of a histogram.
Both of them are normalized since KL divergence applies to probability distributions.
The higher the value of $D_{KL}$, the more separable those two distributions are, in terms of decoupling the two distributions through an angular threshold with minimum false positives or negatives.
Yet, this metric does not account for topological distances, but quantifies the bin-wise difference between the distributions according to the individual values of $x$ (histogram bins).
As an instance, in Fig.~\ref{fig:Hist}, the left histogram displays two distributions with lower overlap than the right one, also leading to a lower $D_{KL}$ value.
However, observing the horizontal axis, we notice that the right one presents higher angular distance between the two distributions that is not captured by $D_{KL}$.

\subsubsection{Wasserstein distance~\cite{kantorovich2006translocation}} 
also known as Earth Mover's distance ($D_{EM}$), is designed to cover the aforementioned characteristic.
Empirically, given two distributions thought as two masses of soil, $D_{EM}$ measures the least amount of work required to match the one mass to the other.
Thus, the idea of ground distance is introduced, measuring the cost for moving samples of the histogram among its horizontal axis.
In our case, the ground distance corresponds to the embeddings' angles in degrees.
Thus, the higher the $D_{EM}$, the more distant the two distributions are among their horizontal axis, highlighting topological differences between the positive and negative pairs' angular distributions.
In Fig.~\ref{fig:Hist}, $D_{EM}$ displays a higher value in the right histogram, thus capturing angular discrimination, in terms of quantifying the existing margin between the two distributions.
In that sense, high $D_{EM}$ values denote a wider range of available threshold values for distinguishing between positive and negative sample pairs.

\begin{figure*}[h]
    \centering
    \begin{subfigure}[b]{0.49\textwidth}
		\centering    	
    	\begin{subfigure}[b]{0.493\textwidth}
        	\centering
        	\includegraphics[width=\textwidth]{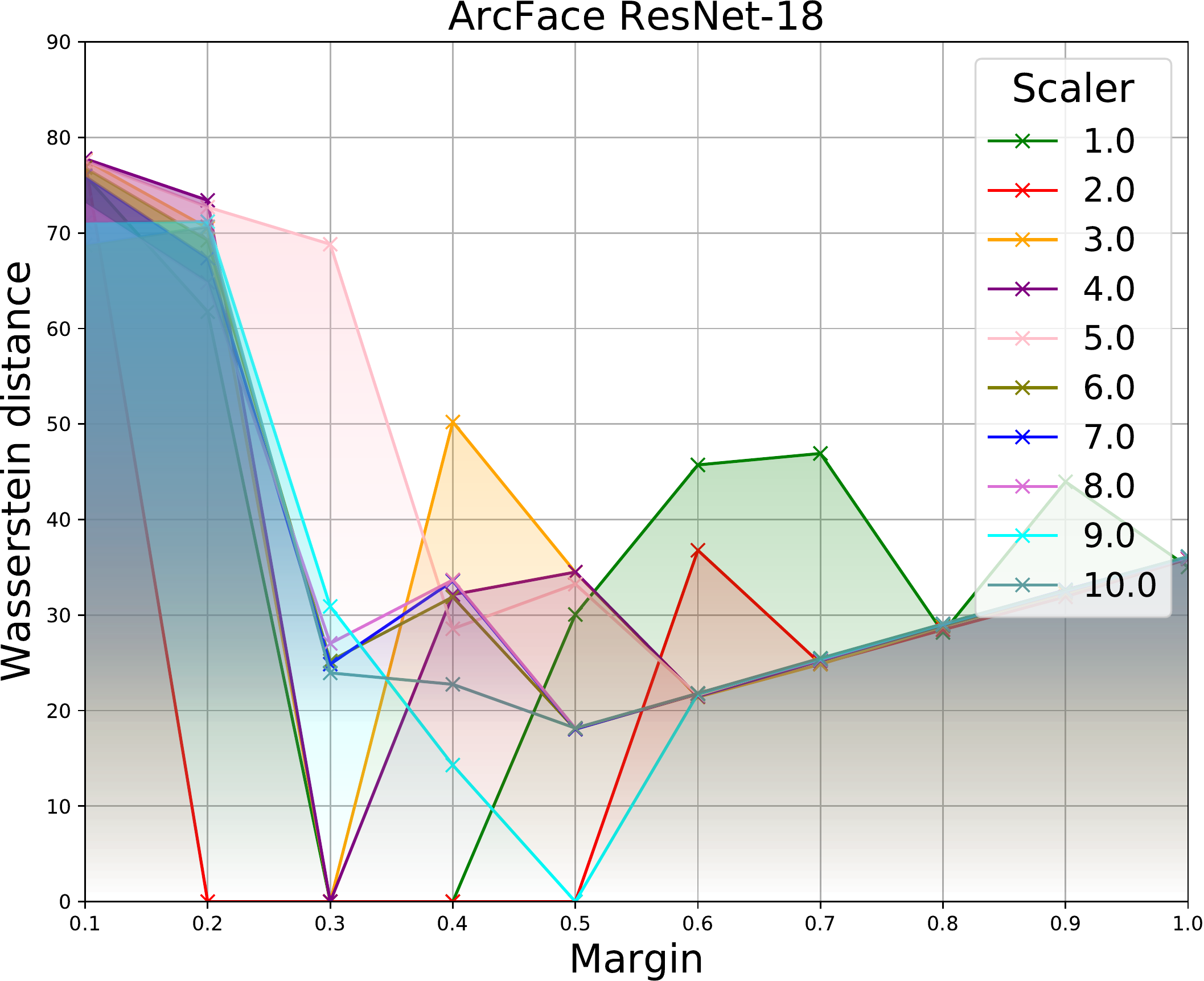}       
    	\end{subfigure}
    	\begin{subfigure}[b]{0.493\textwidth}
        	\centering
        	\includegraphics[width=\textwidth]{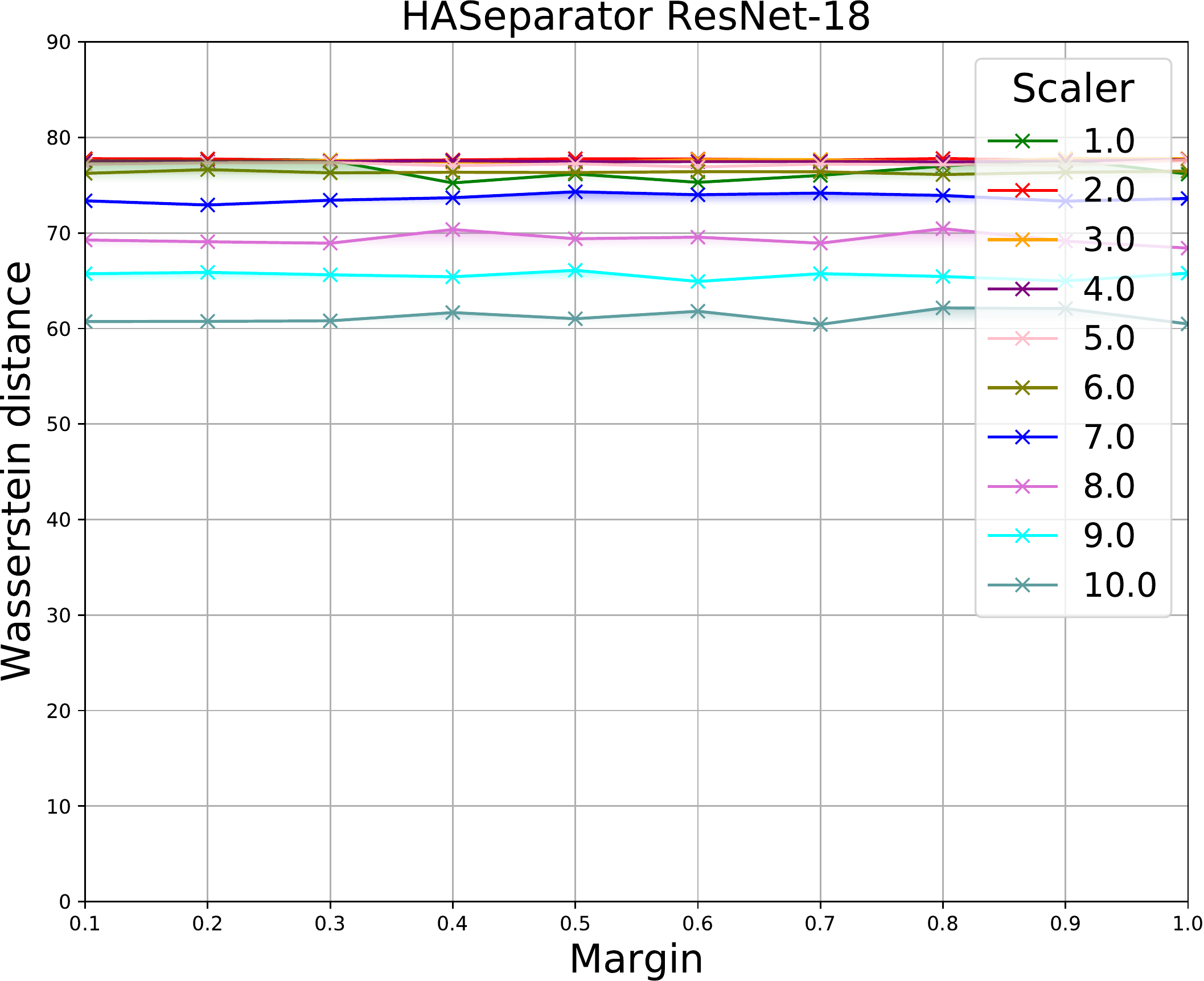}          
    	\end{subfigure}
    	\caption[]{ResNet-18 on SVHN}
    	\label{fig:sub5a}
    \end{subfigure}
    \centering
    \begin{subfigure}[b]{0.49\textwidth}
		\centering    	
    	\begin{subfigure}[b]{0.493\textwidth}
        	\centering
        	\includegraphics[width=\textwidth]{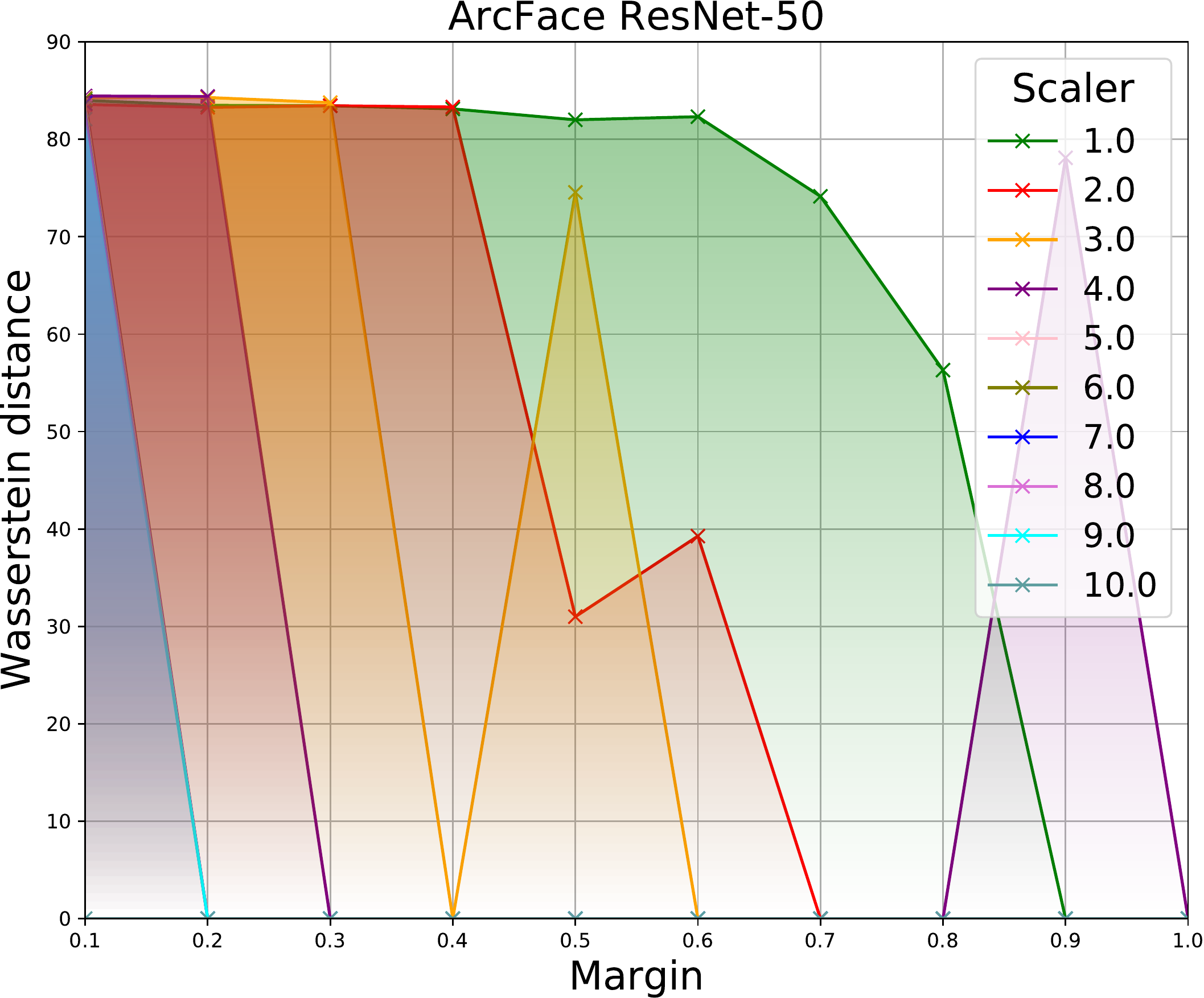}       
    	\end{subfigure}
    	\begin{subfigure}[b]{0.493\textwidth}
        	\centering
        	\includegraphics[width=\textwidth]{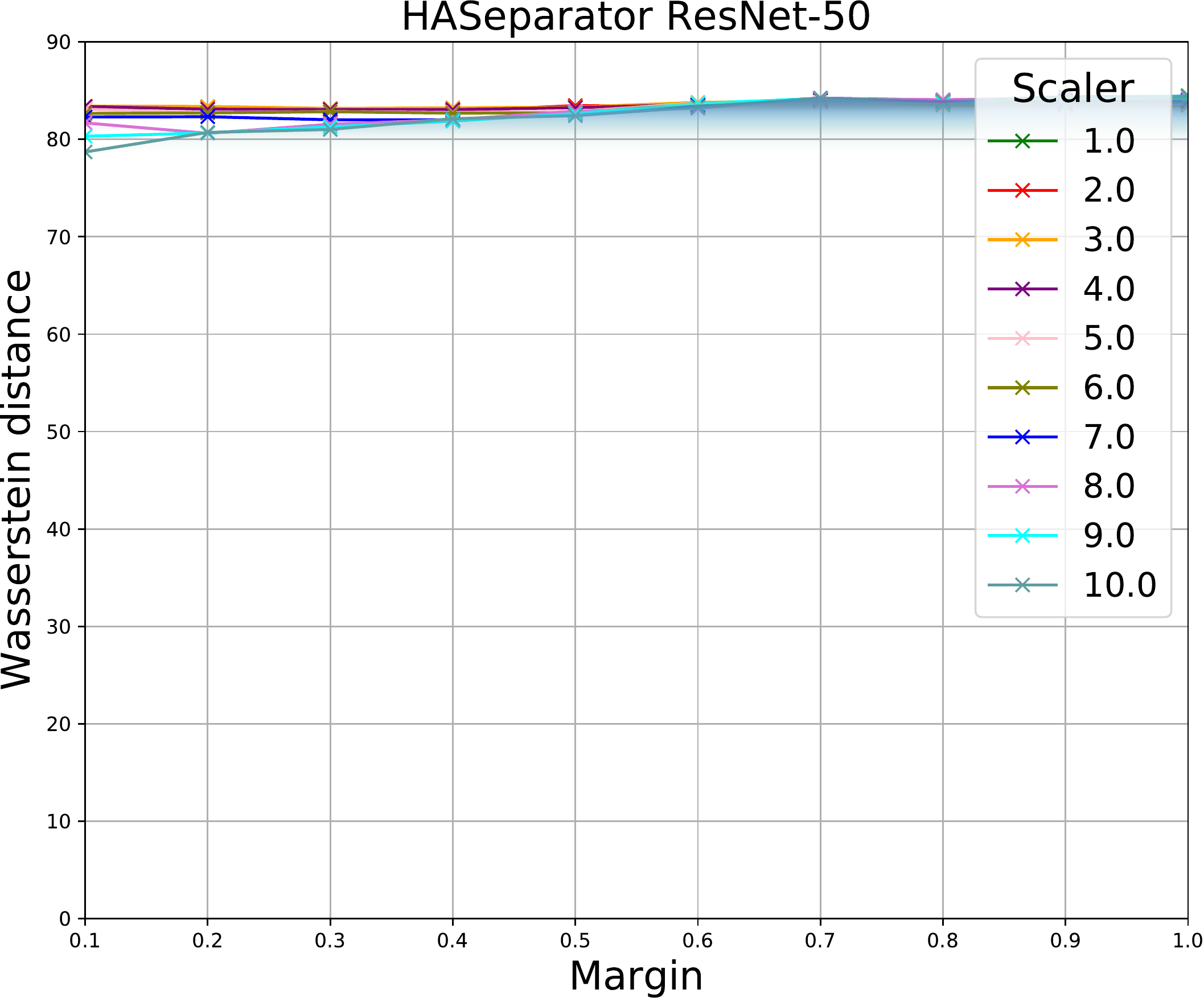}          
    	\end{subfigure}
    	\caption[]{ResNet-50 on SVHN}
    	\label{fig:sub5b}
    \end{subfigure}
    \caption[]
    {\small The obtained Wasserstein distances ($D_{EM}$) between the histograms of positive and negative pairs' angular distributions. Graphs compare the outputs of \textit{ArcFace} and the proposed \textit{HASeparator} for different values of scaler $\sigma$ and margin parameter $m$, employing ResNet-18 and ResNet-50 on SVHN.}
    \label{fig:5}
\end{figure*}
\begin{figure*}[h]
    \centering
    \begin{subfigure}[b]{0.48\textwidth}
        \centering
        \includegraphics[width=0.63\textwidth]{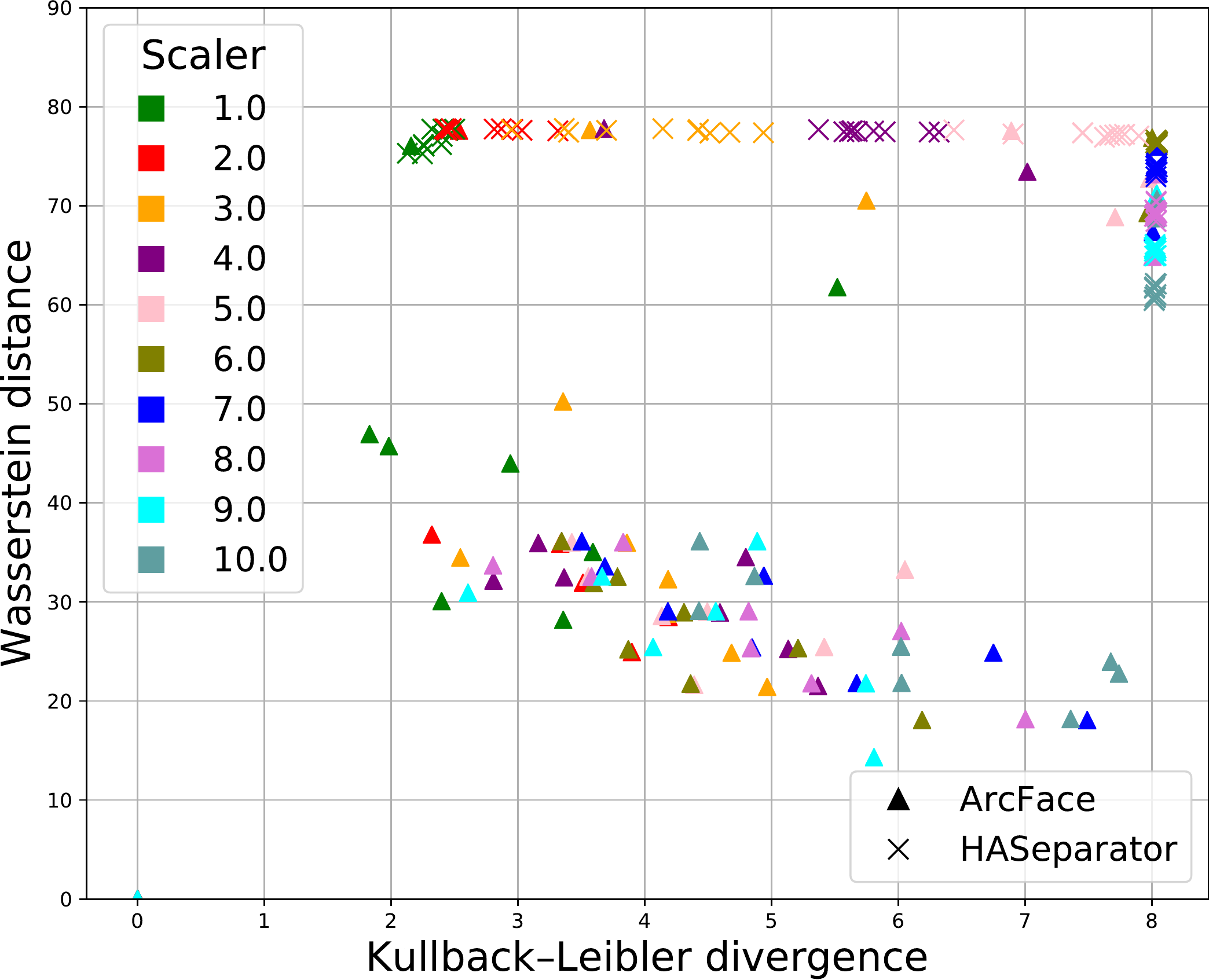}          
    	\caption[]{ResNet-18 on SVHN}
    	\label{fig:sub6a}
    \end{subfigure}
    \centering
    \begin{subfigure}[b]{0.48\textwidth}
		\centering    	
        \includegraphics[width=0.63\textwidth]{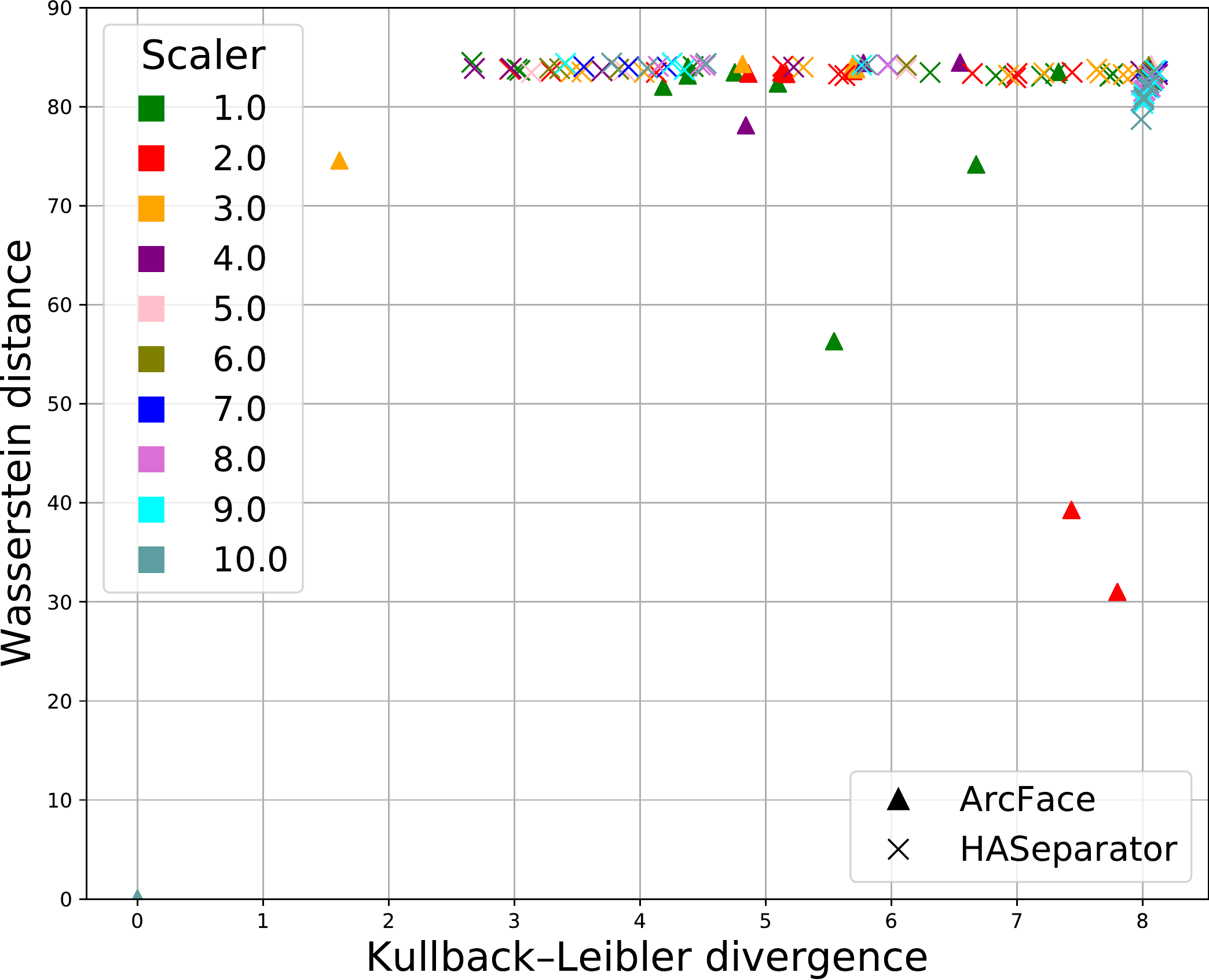}          
    	\caption[]{ResNet-50 on SVHN}
    	\label{fig:sub6b}
    \end{subfigure}
    \caption[]
    {\small Visual assessment of different configurations and the general behavior of \textit{ArcFace} and \textit{HASeparator} through $D_{KL}$ and $D_{EM}$.}
    \label{fig:6}
\end{figure*}

\subsection{Discriminative Feature Learning on CIFAR}\label{DicsFeat}
\textit{HASeparator}, similarly to \textit{ArcFace}, includes a scale factor $\sigma$ that defines the radius of the hypersphere, on which the feature embeddings are located, as well as a margin parameter $m$.
Those two hyperparameters need to be configured properly in order to achieve feature discrimination and at least benchmark classification accuracy. 
Since $m$ is applied to cosine values in both methods, it was sampled from the range $[0.1, 1.0]$ with a step of $0.1$.
Empirically, we observed that for large values of $\sigma$, \textit{i.e.} $16$ or $32$, the $D_{EM}$ drastically decreases.
Hence, we sought values of $\sigma$ in $[1,10]$ with a step of $1$.
The search was performed both for \textit{ArcFace} and \textit{HASeparator} on C10 and C100, once with ResNet-18 and then with \mbox{ResNet-50}.
After each trial, we kept the $D_{EM}$ values among the negative and positive pairs of the testing data to indicate the discrimination performance.
The produced results are summarized in Fig.~\ref{fig:3}.

Beginning with the more simple challenge of C10, see Figs.~\ref{fig:sub3a} and~\ref{fig:sub3c}, we realize that as long as $m$ remains relatively small, indicating a more loose discrimination condition, both methods present quite similar $D_{EM}$ values.
With the increase of $m$, \textit{ArcFace} collapses while \textit{HASeparator} retains its discrimination performance.
For lower values of $\sigma$, this collapse escalates more gradually, especially in the case of the bigger ResNet-50 in Fig.~\ref{fig:sub3c}.
Moreover, in both methods, we notice that for higher values of $\sigma$, $D_{EM}$ starts from lower values and gradually improves reaching the higher ones.
All the above, are verified in a straight-forward manner for the case of the C100, as well.
As an overall result, we can ascertain the more robust nature of \textit{HASeparator}.

Considering the topological properties of $D_{EM}$, it can be overwhelmed by extreme differences in the horizontal histogram axis, failing to capture the effect of per bin overlap that measures the separation capacity of two distributions.
To that end, we also calculate $D_{KL}$ of each configuration, discovering a trade-off relation between those metrics.
More specifically, we place both of them on an orthogonal system with the horizontal axis corresponding to $D_{KL}$, and the vertical one to $D_{EM}$, thus leading to the diagrams in Fig.~\ref{fig:4}.
The illustrations compare the overall permanence of the proposed approach with \textit{ArcFace}, and, they strongly denote the insensitivity to hyperparameters tuning for \textit{HASeparator}, producing a large amount of similar results under a wide range of configuration variations.
This credibility is also established under different number of classes and architectures.
Furthermore, focusing on the upper right corner of each diagram, \textit{HASeparator} offers superior discrimination results, except for the case of ResNet-50 on C100 (Fig.~\ref{fig:sub4d}), where no apparent winner is discerned.

To further study the efficiency of \textit{HASeparator}, we choose the top-2 configurations of each method for both architectures and datasets, and we present their analytic performance metrics in Table~\ref{table:Top2}.
Our choice is based on finding the more distant points in Fig.~\ref{fig:4} and validating that the classification accuracy meets or even surpasses the baseline softmax one.
In C10, \textit{HASeparator} offers in any case increased angular distance between the positive and negative pairs' distributions, meanwhile sustaining their divergence at competitive levels.
Considering the interpretation of $D_{EM}$ and $D_{KL}$ in Section~\ref{metrics}, \textit{HASeparator} increases the available range for placing an angular threshold that decouples the two distributions with similar separation capacities.
In ResNet-18 on C100, the best configuration of \textit{ArcFace} outperforms the second best of ours, though remaining inferior from the best one.
In addition, ResNet-50 on C100 demonstrates a marginal advantage for \textit{ArcFace} with similar $D_{EM}$ values, though higher $D_{KL}$.
Finally, comparing the top-1 against the top-2 of each configuration, we note the trade-off relation of those metrics.

\subsection{Discriminative Feature Learning on SVHN}
We also evaluate feature learning capabilities and insensitivity to different hyperparameter configurations on SVHN.
Thus, we assess the effect of $\sigma$ and $m$ both on \textit{ArcFace} and the proposed \textit{HASeparator} in the same value ranges with CIFAR.
The obtained $D_{EM}$ values are displayed in Fig.~\ref{fig:5}, where we can ascertain a similar behavior with CIFAR.
In specific, our method displays more stable $D_{EM}$ values for various configurations of $\sigma$ and $m$.
Moreover, \textit{ArcFace} notably collapses for large values of $m$.
Fig.~\ref{fig:6} further demonstrates the stability of \textit{HASeparator} on SVHN, producing a larger amount of competitive configurations.
On the contrary, \textit{ArcFace} seems to perform sufficiently in specific hyperparameter layouts.

\begin{table}
\centering
\caption{\small Top-2 configurations of \textit{ArcFace} and \textit{HASeparator} for ResNet-18 and ResNet-50 on SVHN.}
\label{table:Top2svhn}
\resizebox{\linewidth}{!}{%
\renewcommand{\arraystretch}{1.2}
\begin{tabular}{cc|c|c|c|c|c|c|c}
\multirow{3}{*}{} & \multicolumn{8}{c}{\textbf{SVHN}} \\
 & \multicolumn{4}{c|}{\textbf{ResNet-18}} & \multicolumn{4}{c}{\textbf{ResNet-50}} \\
  \cline{2-9}
  & \multicolumn{2}{|c|}{\textit{ArcFace}} & \multicolumn{2}{c|}{\textit{HASeparator}} 
  & \multicolumn{2}{c|}{\textit{ArcFace}} & \multicolumn{2}{c|}{\textit{HASeparator}} \\
\hline
\multicolumn{1}{c|}{$\sigma$} & $4.0$ & $5.0$ & $3.0$ & $5.0$ &
$4.0$ & $5.0$ & $9.0$ & $5.0$ \\
\multicolumn{1}{c|}{$m$} & $0.1$ & $0.1$ & $0.9$ & $1.0$ &
$0.1$ & $0.1$ & $1.0$ & $0.6$ \\
\hline
\multicolumn{1}{c|}{$D_{KL}$} & $3.679$ & $\textbf{6.891}$ & $3.366$ & $6.439$ &
$6.545$ & $8.067$ & $2.660$ & $\textbf{8.099}$ \\
\multicolumn{1}{c|}{$D_{EM}$} & $77.78$ & $77.54$ & $\textbf{77.80}$ & $77.64$ &
$84.44$ & $84.30$ & $\textbf{84.47}$ & $83.37$ \\
\end{tabular}}
\end{table}
Finally, we collect for each approach the top-2 configurations among the two different architectures, as shown in Table~\ref{table:Top2svhn}.
Both approaches obtain very close top-2 discrimination performance on SVHN, rendering it difficult to discern a winner on this dataset. 
Additionally, they achieve about the same classification accuracy at $96.20\%$, while the corresponding performance of softmax loss on SVHN is $94.50\%$.

\section{Conclusion}
To sum up, the paper at hand introduces an innovative method for learning discriminative features with CNNs, based on the optimization of the angular distances between each feature embedding and the separation hyperplanes of its target class.
The proposed approach achieves state-of-the-art results on image classification benchmarks competing \textit{ArcFace}, forming the most acknowledged approach in the field.
Nonetheless, \textit{HASeparator} leads to a more stable performance that remains invariant to hyperparameters tuning to a great extend, as shown by our experimental evaluation in Figs.~\ref{fig:4} and~\ref{fig:6}.
As part of our future work, we plan to test \textit{HASeparator} in the specific task of face verification to further discover its feature learning capacity on this open-set challenge.
In addition, the proposed method could be probably proved beneficial to the problem of uncertainty between estimations under noisy data. 

\section*{Acknowledgment}
This work was supported by Google's TensorFlow Research Cloud and Google's Research Credits programme.

\bibliographystyle{IEEEtran}
\bibliography{IEEEabrv,root}

\begin{thebibliography}{10}
\providecommand{\url}[1]{#1}
\csname url@samestyle\endcsname
\providecommand{\newblock}{\relax}
\providecommand{\bibinfo}[2]{#2}
\providecommand{\BIBentrySTDinterwordspacing}{\spaceskip=0pt\relax}
\providecommand{\BIBentryALTinterwordstretchfactor}{4}
\providecommand{\BIBentryALTinterwordspacing}{\spaceskip=\fontdimen2\font plus
\BIBentryALTinterwordstretchfactor\fontdimen3\font minus
  \fontdimen4\font\relax}
\providecommand{\BIBforeignlanguage}[2]{{%
\expandafter\ifx\csname l@#1\endcsname\relax
\typeout{** WARNING: IEEEtran.bst: No hyphenation pattern has been}%
\typeout{** loaded for the language `#1'. Using the pattern for}%
\typeout{** the default language instead.}%
\else
\language=\csname l@#1\endcsname
\fi
#2}}
\providecommand{\BIBdecl}{\relax}
\BIBdecl

\bibitem{liu2017survey}
W.~Liu, Z.~Wang, X.~Liu, N.~Zeng, Y.~Liu, and F.~E. Alsaadi, ``A survey of deep
  neural network architectures and their applications,'' \emph{Neurocomputing},
  vol. 234, pp. 11--26, 2017.

\bibitem{santavas2020attention}
N.~Santavas, I.~Kansizoglou, L.~Bampis, E.~Karakasis, and A.~Gasteratos,
  ``Attention! a lightweight 2d hand pose estimation approach,'' \emph{arXiv
  preprint arXiv:2001.08047}, 2020.

\bibitem{arandjelovic2016netvlad}
R.~Arandjelovic, P.~Gronat, A.~Torii, T.~Pajdla, and J.~Sivic, ``Netvlad: Cnn
  architecture for weakly supervised place recognition,'' in \emph{Proc. IEEE
  Conf. Comput. Vis. Pattern Recognit.}, 2016, pp. 5297--5307.

\bibitem{rawat2017deep}
W.~Rawat and Z.~Wang, ``Deep convolutional neural networks for image
  classification: A comprehensive review,'' \emph{Neural Computation}, vol.~29,
  pp. 2352--2449, 2017.

\bibitem{kansizoglou2019active}
I.~Kansizoglou, L.~Bampis, and A.~Gasteratos, ``An active learning paradigm for
  online audio-visual emotion recognition,'' \emph{IEEE Trans. Affective
  Comput.}, 2019.

\bibitem{masi2018deep}
I.~Masi, Y.~Wu, T.~Hassner, and P.~Natarajan, ``Deep face recognition: A
  survey,'' in \emph{Proc. 31st Conf. Graph., Patterns, Imag.}, 2018, pp.
  471--478.

\bibitem{deng2019arcface}
J.~Deng, J.~Guo, N.~Xue, and S.~Zafeiriou, ``Arcface: Additive angular margin
  loss for deep face recognition,'' in \emph{Proc. IEEE Conf. Comput. Vis.
  Pattern Recognit.}, 2019, pp. 4690--4699.

\bibitem{schroff2015facenet}
F.~Schroff, D.~Kalenichenko, and J.~Philbin, ``Facenet: A unified embedding for
  face recognition and clustering,'' in \emph{Proc. IEEE Conf. Comput. Vis.
  Pattern Recognit.}, 2015, pp. 815--823.

\bibitem{wen2016discriminative}
Y.~Wen, K.~Zhang, Z.~Li, and Y.~Qiao, ``A discriminative feature learning
  approach for deep face recognition,'' in \emph{Proc. Europ. Conf. Comput.
  Vis.}, 2016, pp. 499--515.

\bibitem{zhang2017range}
X.~Zhang, Z.~Fang, Y.~Wen, Z.~Li, and Y.~Qiao, ``Range loss for deep face
  recognition with long-tailed training data,'' in \emph{Proc. IEEE Conf.
  Comput. Vis. Pattern Recognit.}, 2017, pp. 5409--5418.

\bibitem{cai2018island}
J.~Cai, Z.~Meng, A.~S. Khan, Z.~Li, J.~O’Reilly, and Y.~Tong, ``Island loss
  for learning discriminative features in facial expression recognition,'' in
  \emph{Proc. 13th IEEE Int. Conf. Autom. Face Gesture Recognit.}, 2018, pp.
  302--309.

\bibitem{wang2017normface}
F.~Wang, X.~Xiang, J.~Cheng, and A.~L. Yuille, ``Normface: L2 hypersphere
  embedding for face verification,'' in \emph{Proc. 25th ACM Int. Conf.
  Multimedia}, 2017, pp. 1041--1049.

\bibitem{ranjan2017l2}
R.~Ranjan, C.~D. Castillo, and R.~Chellappa, ``L2-constrained softmax loss for
  discriminative face verification,'' \emph{arXiv preprint arXiv:1703.09507},
  2017.

\bibitem{liu2017sphereface}
W.~Liu, Y.~Wen, Z.~Yu, M.~Li, B.~Raj, and L.~Song, ``Sphereface: Deep
  hypersphere embedding for face recognition,'' in \emph{Proc. IEEE Conf.
  Comput. Vis. Pattern Recognit.}, 2017, pp. 212--220.

\bibitem{wang2018cosface}
H.~Wang, Y.~Wang, Z.~Zhou, X.~Ji, D.~Gong, J.~Zhou, Z.~Li, and W.~Liu,
  ``Cosface: Large margin cosine loss for deep face recognition,'' in
  \emph{Proc. IEEE Conf. Comput. Vis. Pattern Recognit.}, 2018, pp. 5265--5274.

\bibitem{maaten2008visualizing}
L.~v.~d. Maaten and G.~Hinton, ``Visualizing data using t-sne,'' \emph{J. Mach.
  Learn. Res.}, vol.~9, pp. 2579--2605, 2008.

\bibitem{taigman2014deepface}
Y.~Taigman, M.~Yang, M.~Ranzato, and L.~Wolf, ``Deepface: Closing the gap to
  human-level performance in face verification,'' in \emph{Proc. IEEE Conf.
  Comput. Vis. Pattern Recognit.}, 2014, pp. 1701--1708.

\bibitem{sun2014deep}
Y.~Sun, X.~Wang, and X.~Tang, ``Deep learning face representation from
  predicting 10,000 classes,'' in \emph{Proc. IEEE Conf. Comput. Vis. Pattern
  Recognit.}, 2014, pp. 1891--1898.

\bibitem{huang2008labeled}
G.~B. Huang, M.~Mattar, T.~Berg, and E.~Learned-Miller, ``Labeled faces in the
  wild: A database forstudying face recognition in unconstrained
  environments,'' 2008.

\bibitem{kemelmacher2016megaface}
I.~Kemelmacher-Shlizerman, S.~M. Seitz, D.~Miller, and E.~Brossard, ``The
  megaface benchmark: 1 million faces for recognition at scale,'' in
  \emph{Proc. IEEE Conf. Comput. Vis. Pattern Recognit.}, 2016, pp. 4873--4882.

\bibitem{deng2017marginal}
J.~Deng, Y.~Zhou, and S.~Zafeiriou, ``Marginal loss for deep face
  recognition,'' in \emph{Proc. IEEE Conf. Comput. Vis. Pattern Recognit.
  Workshops}, 2017, pp. 60--68.

\bibitem{liu2016large}
W.~Liu, Y.~Wen, Z.~Yu, and M.~Yang, ``Large-margin softmax loss for
  convolutional neural networks.'' in \emph{Proc. Int. Conf. Mach. Learn.},
  vol.~2, 2016, p.~7.

\bibitem{zheng2015triangular}
L.~Zheng, K.~Idrissi, C.~Garcia, S.~Duffner, and A.~Baskurt, ``Triangular
  similarity metric learning for face verification,'' in \emph{Proc. 11th IEEE
  Int. Conf. Workshops Autom. Face Gesture Recognit.}, vol.~1, 2015, pp. 1--7.

\bibitem{kansizoglou2020deep}
I.~Kansizoglou, L.~Bampis, and A.~Gasteratos, ``Deep feature space: A
  geometrical perspective,'' \emph{arXiv preprint arXiv:2007.00062}, 2020.

\bibitem{cortes1995support}
C.~Cortes and V.~Vapnik, ``Support-vector networks,'' \emph{Mach. Learn.},
  vol.~20, pp. 273--297, 1995.

\bibitem{einstein1916foundation}
A.~Einstein \emph{et~al.}, ``The foundation of the general theory of
  relativity,'' \emph{Annalen der Physik}, vol.~49, pp. 769--822, 1916.

\bibitem{abadi2016tensorflow}
M.~Abadi, P.~Barham, J.~Chen, Z.~Chen, A.~Davis, J.~Dean, M.~Devin,
  S.~Ghemawat, G.~Irving, M.~Isard \emph{et~al.}, ``Tensorflow: A system for
  large-scale machine learning,'' in \emph{12th Symp. Operating Syst. Des.
  Implement.}, 2016, pp. 265--283.

\bibitem{paszke2017automatic}
A.~Paszke, S.~Gross, S.~Chintala, G.~Chanan, E.~Yang, Z.~DeVito, Z.~Lin,
  A.~Desmaison, L.~Antiga, and A.~Lerer, ``Automatic differentiation in
  pytorch,'' 2017.

\bibitem{he2016deep}
K.~He, X.~Zhang, S.~Ren, and J.~Sun, ``Deep residual learning for image
  recognition,'' in \emph{Proc. IEEE Conf. Comput. Vis. Pattern Recognit.},
  2016, pp. 770--778.

\bibitem{krizhevsky2009learning}
A.~Krizhevsky and G.~Hinton, ``Learning multiple layers of features from tiny
  images,'' \emph{Tech. Report}, 2009.

\bibitem{netzer2011reading}
Y.~Netzer, T.~Wang, A.~Coates, A.~Bissacco, B.~Wu, and A.~Y. Ng, ``Reading
  digits in natural images with unsupervised feature learning,'' \emph{Neural
  Inf. Process. Syst. Workshop}, 2011.

\bibitem{huang2017densely}
G.~Huang, Z.~Liu, L.~Van Der~Maaten, and K.~Q. Weinberger, ``Densely connected
  convolutional networks,'' in \emph{Proc. IEEE Conf. Comput. Vis. Pattern
  Recognit.}, 2017, pp. 4700--4708.

\bibitem{kullback1951information}
S.~Kullback and R.~A. Leibler, ``On information and sufficiency,'' \emph{Ann.
  Math. Statist.}, vol.~22, pp. 79--86, 1951.

\bibitem{kantorovich2006translocation}
L.~V. Kantorovich, ``On the translocation of masses,'' \emph{J. Math. Sci.},
  vol. 133, pp. 1381--1382, 2006.

\end{thebibliography}

\end{document}